\documentclass{article}

\usepackage[preprint]{neurips_2024}
\usepackage[dvipsnames]{xcolor}
\usepackage[utf8]{inputenc} 
\usepackage[T1]{fontenc}    
\usepackage[citecolor=NavyBlue, linkcolor=NavyBlue, colorlinks=True]{hyperref}
\usepackage{wrapfig}
\usepackage{caption}
\usepackage{url}
\usepackage{booktabs}
\usepackage{nicefrac}
\usepackage{microtype}
\usepackage{graphicx}
\usepackage{enumitem}
\usepackage{amsthm, amsmath, amssymb, amsfonts}
\usepackage[capitalize]{cleveref}
\usepackage{mathtools}

\usepackage{mathmacros}
\newcommand{\x}{x}
\newcommand{\y}{y}

\newcommand{\lr}{\eta}

\newcommand{\eps}{\epsilon}
\newcommand{\lam}{\lambda}
\renewcommand{\a}{a}
\renewcommand{\b}{b}
\newcommand{\g}{g}

\newcommand{\p}{p}
\newcommand{\hmat}{H}  

\newcommand{\scale}{\sigma}
\renewcommand{\v}{v} \newcommand{\w}{v_{\perp}}
\DeclareMathOperator{\diag}{diag}
\newcommand{\param}{w}
\newcommand{\CDAT}{CDAT} 
\newcommand{\cdat}{Curvature Dynamics Aware Tuning} 

\let\originalparagraph\paragraph
\renewcommand{\paragraph}[2][.]{\originalparagraph{#2#1}}

\title{Stepping on the Edge: \\ 
Curvature Aware Learning Rate Tuners} \author{ Vincent Roulet\textsuperscript{*}
\\
Google DeepMind\\
\texttt{\small vroulet@google.com}
\And Atish Agarwala\textsuperscript{*}\\
Google DeepMind\\
\texttt{\small thetish@google.com}
\And Jean-Bastien Grill \\
Google DeepMind\\
\texttt{\small jbgrill@google.com}
\And Grzegorz Swirszcz \\
Google DeepMind\\
\texttt{\small swirszcz@google.com}
\AND Mathieu Blondel \\
Google DeepMind\\
\texttt{\small mblondel@google.com}
\And Fabian Pedregosa \\
Google DeepMind\\
\texttt{\small pedregosa@google.com}
}

\begin{document}

\maketitle

\begin{abstract}
Curvature information -- particularly, the largest eigenvalue of the loss
Hessian, known as the sharpness -- often forms the basis for learning rate
tuners. However, recent work has shown that the curvature information undergoes
complex dynamics during training, going from a phase of increasing sharpness to
eventual stabilization. We analyze the closed-loop feedback effect between
learning rate tuning and curvature. We find that classical learning rate tuners
may yield greater one-step loss reduction, yet they ultimately underperform in
the long term when compared to constant learning rates in the full batch regime.
These models break the stabilization of the sharpness, which we explain using a
simplified model of the joint dynamics of the learning rate and the curvature.
To further investigate these effects, we introduce a new learning rate tuning
method, Curvature Dynamics Aware Tuning (CDAT), which prioritizes long term
curvature stabilization over instantaneous progress on the objective. In the
full batch regime, CDAT shows behavior akin to prefixed warm-up schedules on
deep learning objectives, outperforming tuned constant learning rates. In the
mini batch regime, we observe that stochasticity introduces confounding effects
that explain the previous success of some learning rate tuners at appropriate
batch sizes. Our findings highlight the critical role of understanding the joint
dynamics of the learning rate and curvature, beyond greedy minimization, to
diagnose failures and design effective adaptive learning rate tuners.
\end{abstract}

\section{Introduction}

The learning rate, a.k.a.\ stepsize, is the main hyperparameter controlling the
efficiency and stability of gradient-based training of deep neural networks. The
learning rate is typically adjusted through a predetermined schedule -- often
consisting of a warm-up phase, where the learning rate is gradually increased to
a peak, followed by an annealing phase, where it is decreased to
zero~\citep{loshchilov2016sgdr, goyal2017accurate}. Tuning the shape of the
schedule (warm-up time, peak learning rate, decay scale and shape) is essential
for good performance. Despite recent efforts to understand their effectiveness,
the optimal shape of these schedules remains an area of active
research~\citep{liu2019variance, shi2020learning}. The cost of tuning these
schedules has led to interest in automatic selection of these hyperparameters
with \emph{learning rate tuners} - methods which aim to automatically adjust the
learning rate through training. 

These methods have roots in traditional
optimization theory, including inexact linesearch with Armijo-Goldstein
criterion~\citep{armijo1966minimization, nocedal1999numerical} and Polyak
stepsizes~\citep{polyak1964some}, which select the learning rate via estimates
of the gap to optimality of the objective.
The Armijo-Goldstein criterion is a crucial component of popular full-batch
convex optimizers, such as L-BFGS \citep{liu1989limited}. Recent efforts have
adapted linesearches to stochastic optimization, with some partial empirical
successes and with some approaches offering convergence
guarantees~\citep{vaswani2019painless, galli2023don, mutschler2020parabolic}.
Similar efforts have been made for Polyak stepsizes
~\citep{loizou2021stochastic, berrada2020training}, in addition to new methods
which combine distance to optimality with online learning convergence bounds
~\citep{defazio2023learning, cutkosky2023mechanic, mishchenko2023prodigy,
ivgi2023dog}.

Classically-inspired methods, however, have generally struggled to gain traction
in deep learning. This is partly due to their design, which prioritizes convex,
Lipschitz-continuous, and/or smooth (Lipschitz-continuous gradients) objectives.
In contrast, the loss landscape of deep networks is known to be
non-convex~\citep{li2018visualizing}, and non-Lipschitz
continuous~\citep{hochreiter2001gradient}. Moreover, non-linear models,
especially neural networks, will commonly undergo dramatic changes in geometry
during training \citep{jastrzkebski2018relation, wu2018sgd, kopitkov2020neural,
jastrzebski2020break}. In particular, most models undergo a phase of
\emph{progressive sharpening} - where the sharpness, the largest eigenvalue of
the Hessian, increases during training~\citep{cohen2021gradient}. These
potentially detrimental effects are mitigated by non-linear stabilization
arising from the discreteness of the dynamics -- namely, the \emph{edge of
stability} (EOS) phenomenon~\citep{cohen2021gradient}. This causes large Hessian
eigenvalues to stabilize at the critical value for a given learning rate in an
equivalent smooth setting (for example, max Hessian eigenvalue stabilizes at
$\lam_{\max} =  2/\lr$ for learning rate $\lr$) \cite{cohen2021gradient,
cohen2022adaptive}. \citet{gilmer2021loss} considered EOS stabilization as a
leading candidate for the necessity of the warm-up procedure; as the learning
rate $\lr$ increases, $\lam_{\max}$ is effectively annealed. 

This raises some natural questions. \emph{How do these sharpness dynamics affect
the performance of learning rate tuners? What insights can we gain to design
better tuners for deep learning?} Our work takes a first step at answering these
questions, starting with a study of some classical learning rate tuners: a
linesearch ensuring sufficient decrease and an approximately greedy method that
minimizes a quadratic approximation of the objective. Specifically, we find the
following.
\begin{itemize}[nosep, leftmargin=*]
    \item We empirically observe that classical learning rate tuners
    qualitatively underperform their constant learning rate counterparts across
    several deep learning benchmarks, in the full batch regime, for which these
    methods were originally designed.
    \item Our empirical analysis of curvature dynamics reveals that classical
    learning rate tuners generally undershoot the edge of stability. This
    undershooting creates a snowball effect of ever-increasing sharpness and
    ever-decreasing learning rates.
    \item We propose a theoretical model that effectively captures these
    empirically observed failures.
\end{itemize}
Our analysis suggests that stabilizing the sharpness may be a more important
goal for the long-term success of training, compared to greedily optimizing the
objective. To explore this idea, we propose the \cdat{} (\CDAT{}) method, which
dynamically drives the learning rate to the EOS. In our exploration, we find the
following.
\begin{itemize}[nosep, leftmargin=*]
    \item We observe empirically that the proposed learning rate tuner can
    outperform fine-tuned constant learning rate counterparts in a full batch
    regime. 
    \item We analyze the sharpness dynamics induced by \CDAT{} in these examples
    and observe that the progressive sharpening is mitigated by the tuner,
    increasing learning rates at early times before stabilizing, akin to an
    automatic warm-up schedule.
    \item We propose a theoretical model that clarifies the dynamical mechanisms
    by which \CDAT{} maintains proximity to the EOS, while highlighting the
    limitations of existing models of curvature dynamics.
\end{itemize}
Our work suggests that the design of learning rate tuners benefits from
exploiting curvature stabilization rather than focusing on loss decrease. The
introduction of simple learning rate tuners can also refine our understanding of
sharpness dynamics through feedback loop effects. Additional experiments and
experimental details are presented in~\cref{app:more_exp}
and~\cref{app:exp_details} respectively.

\section{The Interplay Between Learning Rate Tuners and Curvature Dynamics}
\label{sec:classic_tuners}

A leitmotif in the design of learning rate tuners has been to select the
learning rate to ensure a maximal or sufficient decrease of the objective at
each iteration. We focus here on two canonical examples. Polyak stepsizes and
hyper-gradient descent are also briefly examined in \cref{app:more_exp},
~\cref{fig:other_tuners}.

\subsection{Canonical learning rate tuners failures in deep learning}
\label{ssec:classic_tuners_intro}

\begin{figure}[t]
    \centering
    \includegraphics[width=0.48\linewidth]{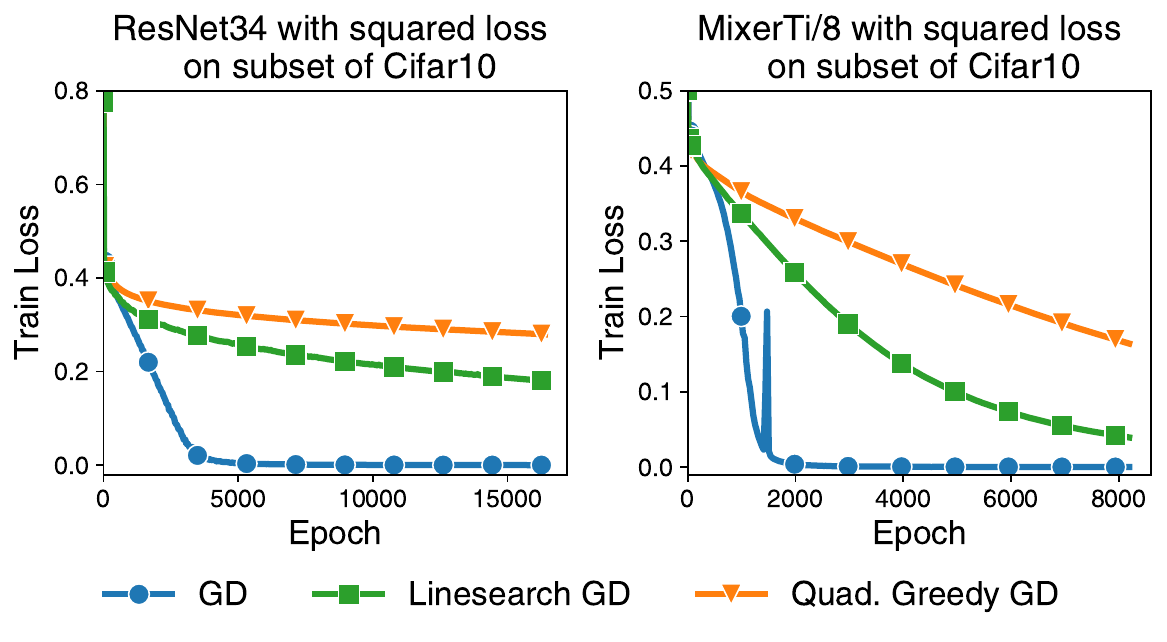}
    \includegraphics[width=0.48\linewidth]{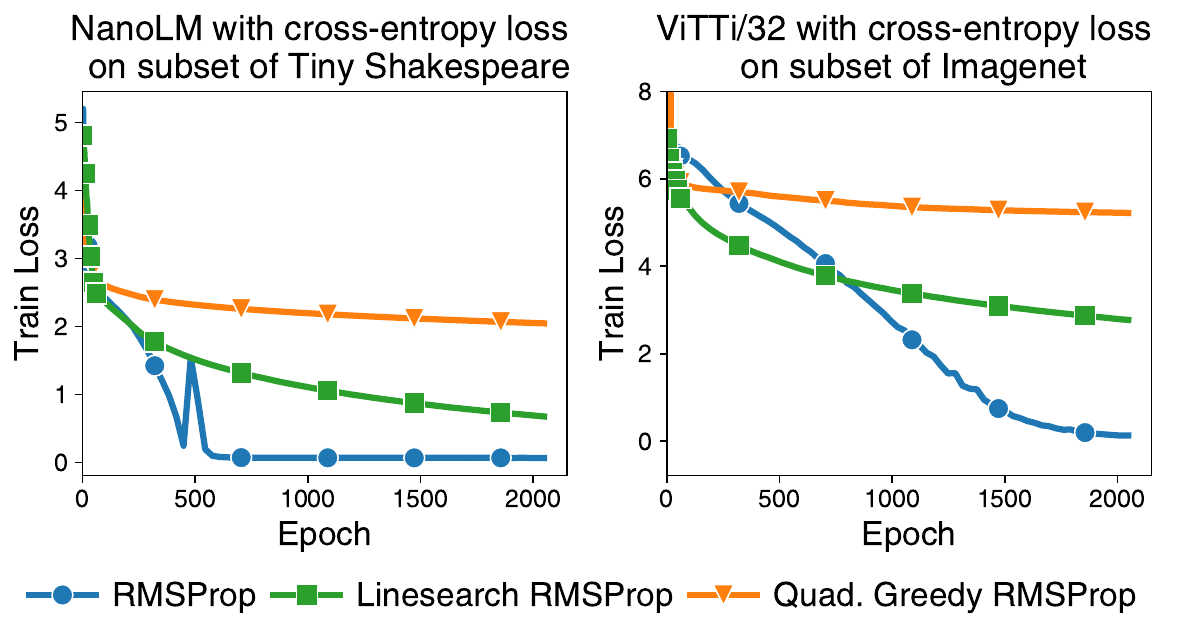}
    \caption{\textbf{Simple learning rates tuners qualitatively underperform their constant learning rate counterparts.}
    Gradient descent or RMSProp with a tuned constant learning rate versus
    self-tuned gradient descent by a linesearch method~\eqref{eq:linesearch}, or
    a quadratically greedy rule~\eqref{eq:quadratically_greedy} on various
    datasets, architectures and losses in a full batch regime. The linesearch
    may perform better at early times but stalls in the long term.}
    \vspace*{-15pt}
    \label{fig:failures_baselines}
\end{figure}

The first classical approach we consider is a \textbf{linesearch} (ls) method
that selects the learning rate $\lr$ such that the objective $f$ satisfies a
certain decrease criterion~\citep{armijo1966minimization, nocedal1999numerical}.
Formally, given current parameters $\param_t$ and an update direction $u_t$, the
learning rate $\lr_t^{\text{ls}}$ is chosen such that
\begin{align}\label{eq:linesearch}
   f(\param_t + \lr_t^{\text{ls}} u_t) 
   \leq f(\param_t) 
   + c\, \lr_t^{\text{ls}} u_t^\top \nabla f(\param_t)\,.
\end{align}
This rule assumes that $u_t$ is a descent direction ($\nabla f(\param_t)^\top
u_t <0$), which ensures the existence of a learning rate
satisfying~\eqref{eq:linesearch}. This holds true for simple Gradient Descent
(GD) or preconditioned variants like RMSProp~\citep{hinton2012divide}. In the
criterion~\eqref{eq:linesearch}, $c$ is usually a small constant set to
$10^{-4}$ or $0$. A valid learning rate is searched with a usual backtracking
linesearch (\cref{app:exp_details}).

The second method we consider involves selecting the learning rate at each
iteration to minimize a quadratic approximation of the objective. Formally, the
objective $f$ at parameters $\param_t$ can be approximated along an update
direction $u_t$ by a quadratic approximation $q_f$ as
\begin{equation}\label{eq:quad_approx}
f(\param_t+\lr u_t) \approx 
q_f(\lr; \param_t, u_t)
\coloneqq
f(\param_t) 
+ \lr \nabla f(\param_t)^\top u_t
+ \frac{1}{2}\lr^2 u_t^\top \nabla^2 f(\param_t) u_t.
\end{equation}
\begin{wrapfigure}[16]{r}{0.4\textwidth}
\centering
\vspace*{-15pt}
\includegraphics[width=0.9\linewidth]{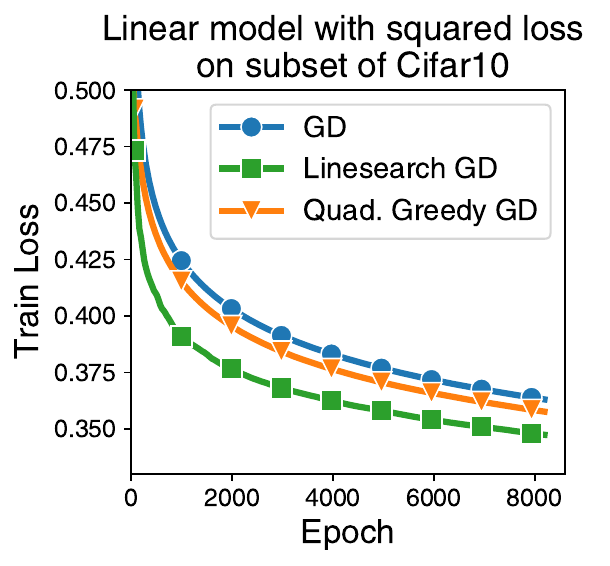}
\captionof{figure}{ 
Classical learning rate tuners can be effective on linear models.
}
\label{fig:linear_classic_baselines}
\end{wrapfigure}
Provided that this quadratic approximation is strongly convex in $\eta$
($u_t^\top \nabla^2 f(\param_t) u_t>0$), the minimum of the quadratic
approximation $q_f(\lr; \param_t, u_t)$ is reached for the \textbf{quadratically
greedy} (qg) learning rate $\lr^{\text{qg}}$ given by
\begin{align}\label{eq:quadratically_greedy}
    \lr_t^{\text{qg}} = \frac{-\nabla f(\param_t)^\top u_t }{u_t^\top \nabla^2 f(\param_t) u_t }\,.
\end{align}
Setting the learning rate by minimizing the quadratic
approximations~\eqref{eq:quadratically_greedy} is a simple intuitive idea
studied for example by~\citet{schaul2013no}, \citet[Section
6.4]{martens2015optimizing}. This approach as well as linesearches are effective
on simple linear problems (\cref{fig:linear_classic_baselines}). While their
rationale originates in non-stochastic optimization, they have been analyzed in
the context of stochastic optimization for deep
learning~\citep{vaswani2019painless, schaul2013no}.

\subsection{Analyzing learning rate tuners through curvature dynamics}

\begin{figure}[t]
    \centering
    \includegraphics[width=\linewidth]{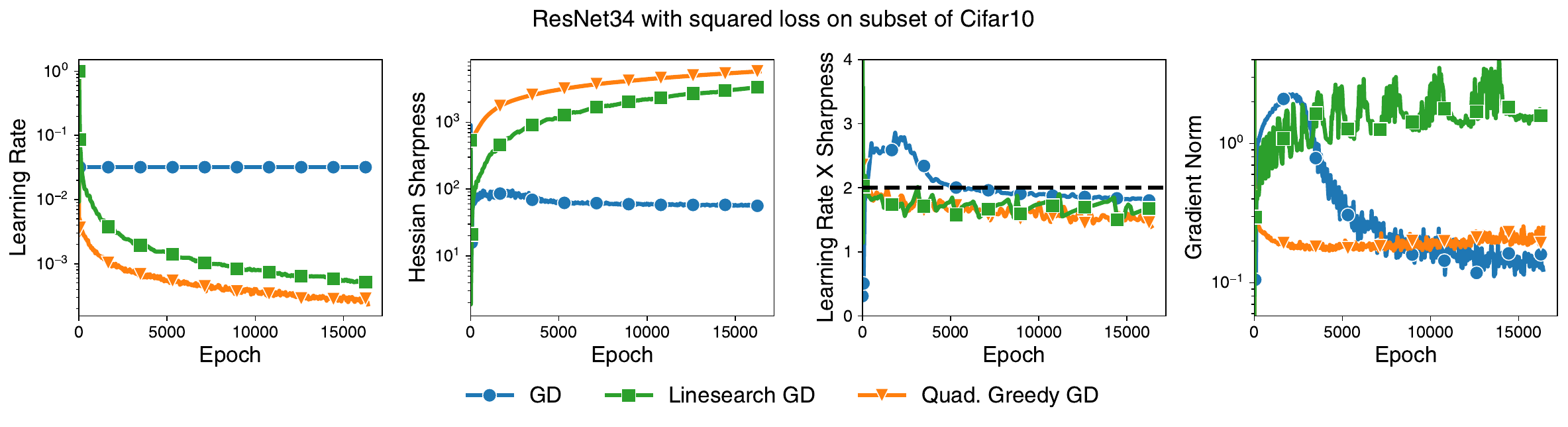}
    \caption{\textbf{Classical learning rate tuners can undershoot the edge of
    stability.} Learning rate, sharpness, their product, and the gradient norm
    evolution of a constant learning rate and learning rate tuners, full batch
    gradient descent. Learning rate decreases by $3$ orders of magnitude for
    tuners ($1$\textsuperscript{st} panel) while sharpness increases
    ($2$\textsuperscript{nd} panel). Their product remains relatively steady,
    just below the edge of stability ($3$\textsuperscript{rd} panel). The
    gradient norm increases by less than a factor of $10$, consistent with slow
    training at late times ($4$\textsuperscript{th} panel). 
    }
    \label{fig:main_sharp_failures}
    \vspace*{-15pt}
\end{figure}

\paragraph{Full batch regime}
We revisit the performance of the learning tuners presented
in~\cref{ssec:classic_tuners_intro} in the full batch regime on deep learning
problems in \cref{fig:failures_baselines}. As demonstrated in
\cref{fig:failures_baselines}, a linesearch~\eqref{eq:linesearch} or the
quadratically greedy rule~\eqref{eq:quadratically_greedy} qualitatively
underperform their constant learning rate counterpart in the deep learning
benchmarks considered. Notably, all these results are obtained despite being in
a full batch regime, for which these methods are originally designed. To
understand the failures of these approaches, we consider several measures
presented in \cref{fig:main_sharp_failures} (see
also~\cref{fig:app_sharp_failures}). 

First, we observe a consistent decrease in the chosen learning rate over time,
spanning several orders of magnitude ($1$\textsuperscript{st} panel of
\cref{fig:main_sharp_failures}). This is surprising, as none of these approaches
explicitly encode a decreasing learning rate mechanism. Specifically, the
linesearch always initiates its search with a guess larger than the previously
selected learning rate (see \cref{app:exp_details} for implementation details).
Decreasing learning rates are theoretically optimal for non-smooth
objectives~\citep{nesterov2018lectures}, such as the ones induced by using the
ReLU activation; however in our example, the gradient norm does not increase
beyond one order of magnitude ($4$\textsuperscript{th} panel
of~\cref{fig:main_sharp_failures}). This suggests both that an increase in
gradient norm is not the primary cause of learning rate decrease, and also
explains why the learning rate decrease is correlated with slower progress on
the training loss.

Following the work of~\citet{cohen2021gradient}, we analyze the dynamics of the
sharpness, that is the largest eigenvalue of the Hessian,
$\lambda_{\max}(\nabla^2 f(\param_t))$. In the $2$\textsuperscript{nd}  panel of
\cref{fig:main_sharp_failures}, we observe that while sharpness stabilizes for
gradient descent, it does not exhibit the same behavior for the considered
learning rate tuners. By plotting the product of the learning rate $\lr_t$ and
the sharpness ($3$\textsuperscript{rd} panel of \cref{fig:main_sharp_failures}),
we find that this product can exceed the stability threshold of $2$, eventually
stabilizing below this threshold for constant learning rate gradient descent. In
contrast, for the learning rate tuners, this product neither surpasses the
stability threshold nor stabilizes around $2$ in the long run. Therefore, these
classical learning rate tuners do not operate at the edge of stability.

From a theoretical perspective, objectives are typically classified as either
smooth or non-smooth. Smooth objectives have gradients that are
Lipschitz-continuous, at least locally around any point. Non-smooth objectives,
on the other hand, may contain points with kinks (non-differentiable points).
However, this taxonomy might not fully capture the curvature dynamics observed
by~\citet{cohen2021gradient, cohen2022adaptive} for constant learning rates, and
in \cref{fig:failures_baselines} for the classical learning rate tuners. In
particular, the concept of smoothness might  not be entirely relevant in the
context of deep learning, where its local estimate (the spectral norm of the
Hessian, also known as sharpness) can continue to increase throughout training.
To push the limits of classical smoothness assumptions, we consider
in~\cref{sec:on_edge} a learning rate tuner that propels the optimizer at the
edge of stability or above, a regime that usual smoothness assumptions would
theoretically prohibit.

\paragraph{Mini-batch regime}
The results presented in~\cref{fig:failures_baselines} in the full batch regime
\emph{do not contradict} previous results from~\citet{vaswani2019painless}, who
studied linesearches for deep learning in the stochastic regime, as illustrated
in~\cref{fig:baselines_stoch}, and previously reported
by~\citet{roulet2023interplay}. We simply point out that the success of
linesearches observed by~\citet{vaswani2019painless} may not be entirely
attributable to the method's original rationale.

The actual success of linesearches in a stochastic regime may instead be
explained by the attenuated progressive sharpening observed in such a regime
\citep{jastrzebski_three_2018, cohen2021gradient, agarwala_high_2024}. Moreover,
linesearches applied to mini-batches tend to select larger learning rates than
they would in a full-batch regime~\citep{mutschler2020parabolic} potentially
allowing them to avoid undershooting the full objective's edge of stability.

\subsection{Theoretical analysis}

\begin{figure}[t]
\centering
\begin{tabular}{ccc}
\includegraphics[height=0.25\linewidth]{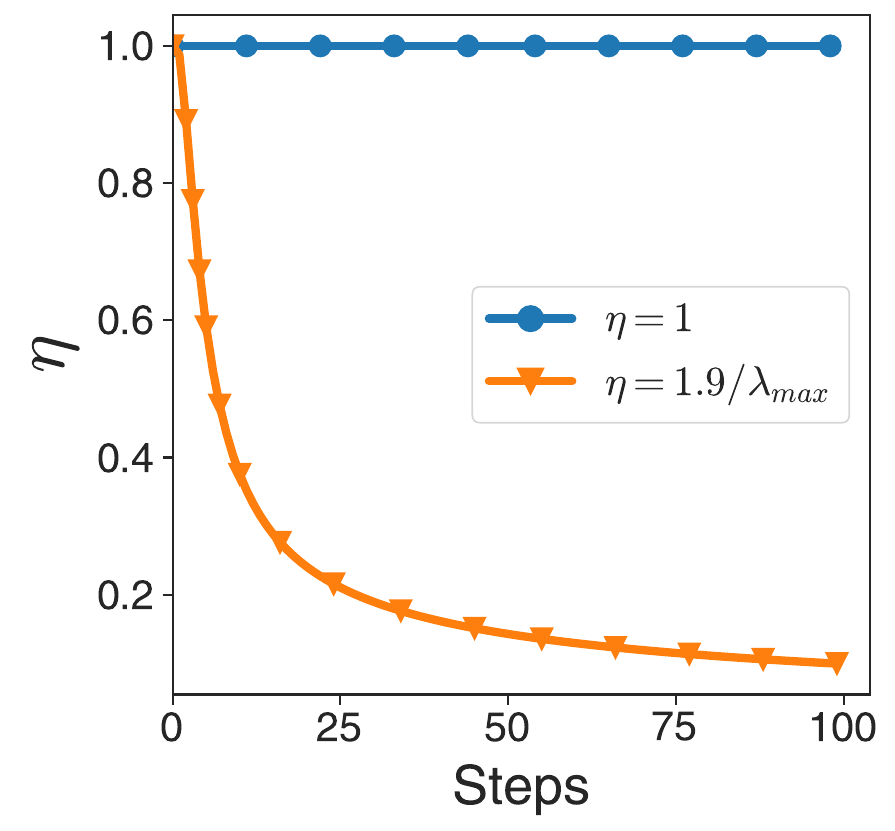} &
\includegraphics[height=0.25\linewidth]{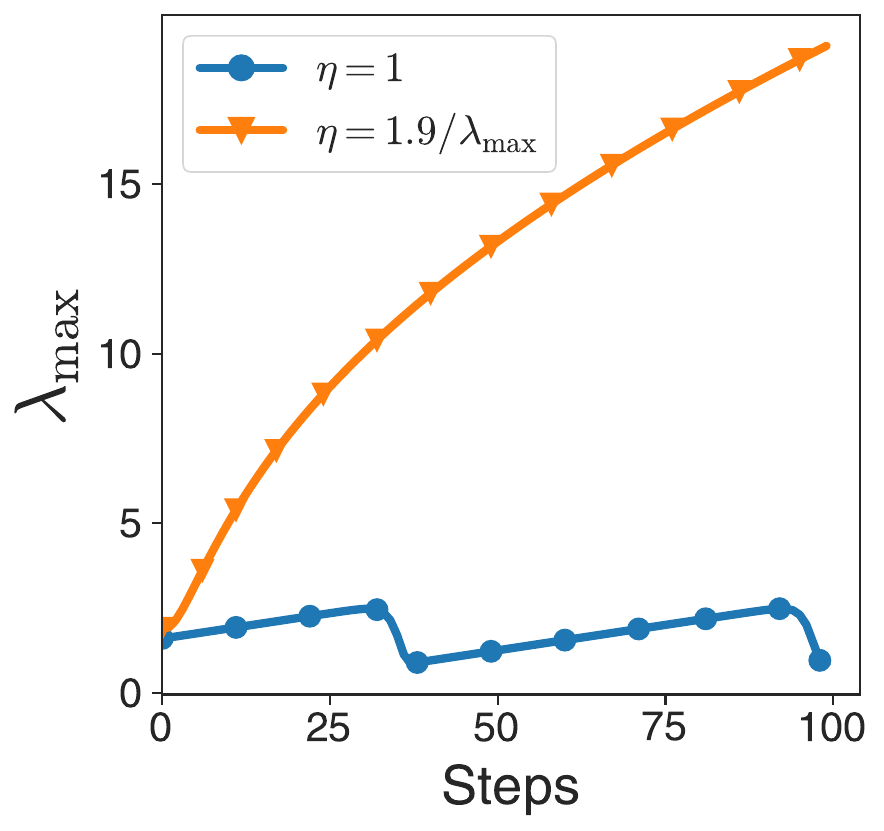} &
\includegraphics[height=0.25\linewidth]{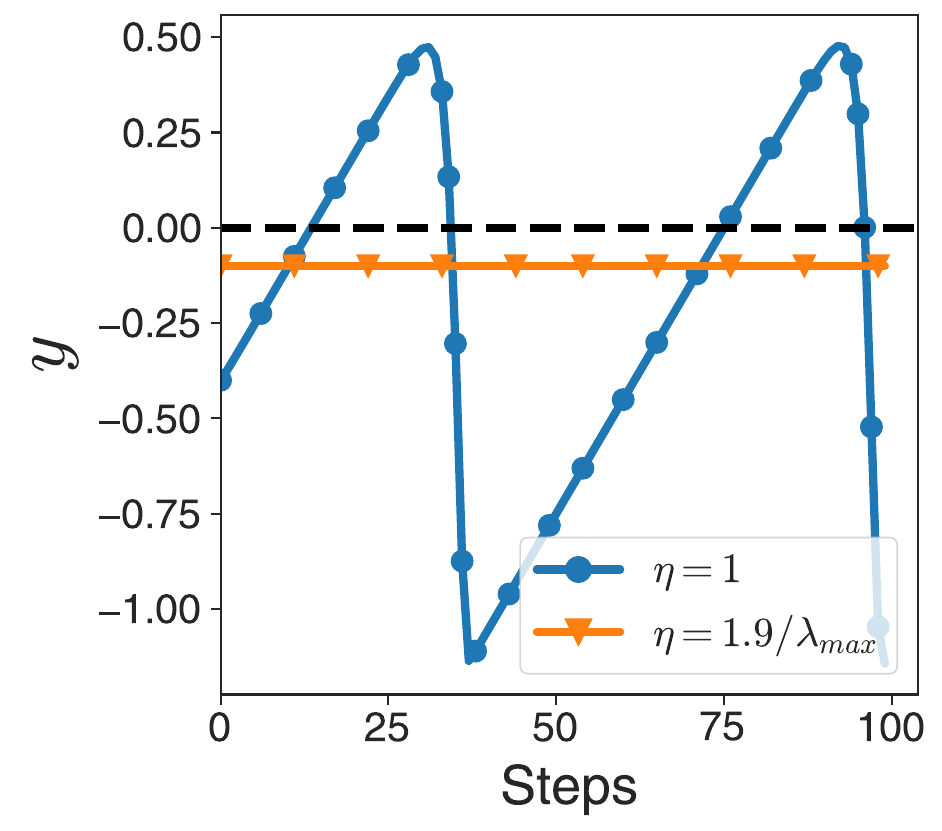}
\end{tabular}
\caption{ \textbf{The poor performance of classical learning rate tuners,
understood in a simplified model.} The dynamics of learning rate $\lr$,
sharpness $\lam_{\max}$, and normalized centered sharpness $\y =
\lr\lam_{\max}-2$ are examined in the simplified model~\eqref{eq:x_l_theory}.
With a constant $\lr$, $\lam_{\max}$ stabilizes and $\y$ oscillates around $0$
(blue). Classical learning rate tuners often quickly equilibrate around $\y_{t}
= -\eps$, which we model using $\eta = 1.9 \lambda_{\max}$ (orange). This
equilibration of $\y$ away from zero prevents stabilization in $\lam_{\max}$,
leading to an increase in $\lam_{\max}$, and a corresponding decrease in $\lr$.
}
\label{fig:armijo_theory}
\vspace*{-15pt}
\end{figure}

\label{sec:theory_part_1}

The sharpening effects can be understood theoretically. Previous work has shown
that the stabilization provided by EOS is due to non-linear interaction between
the component of the gradient in the largest eigendirection, and the dynamics of
the largest eigenvalues themselves \citep{damian2022self,
agarwala_secondorder_2022}. We can use these analyses to understand why there is
no stabilization for some classical learning rate tuners.

We start with the model from \citet{damian2022self}, which focuses on the
dynamics in the largest eigendirection of the Hessian. Given an objective $f$
parameterized by parameters $\param_{t}$, let $\lam_{t}$ be the largest
eigenvalue of the Hessian $\nabla^{2}f(w_t)$, i.e., $\lam_t \coloneqq \lam(w_t)
\coloneqq \lambda_{\max}(\nabla^2 f(w_t))$. Let $\v$ be its normalized
eigenvector; the model assumes slow eigenvector change, so it is treated as a
fixed direction. The joint dynamics of $\lam_{t}$ and the projection
$\x_{t}\coloneqq\v^\top \param_{t}$ can then be written as
\begin{equation}
\x_{t+1} = (1-\lr_{t}\lam_{t})\x_{t},~\lam_{t+1} = \lr_{t}(\a-\b\x_{t}^{2})+\lam_{t}\,.
\label{eq:x_l_theory}
\end{equation}
Here, $\a \coloneqq -\nabla \lam(w)^\top \nabla f(w)$ corresponds to the
instantaneous change of $\lambda$ along the negative gradient (the update
direction), and $\b \coloneqq \|\nabla \lam(w)\|^{2}$ encodes the non-linear
negative feedback between $\x_{t}$ and $\lam_{t}$. Both $a$ and $b$ are
considered constant along iterations. These equations are derived by
\citet{damian2022self} using a Taylor expansion of the iterates combined with a
coupling argument.

In the original model, the learning rate $\lr_{t}$ is also fixed to $\eta$. This
leads to the following dynamics: while $\lr\lam_{t}<2$, the magnitude of
$\x_{t}$ decreases. This, in turn, leads to an increase in $\lam_{t}$.
Eventually, $\lr\lam_{t}>2$ and $|\x_{t}|$ increases. This eventually leads to
the $\b\x_{t}^{2}$ term becoming large, which decreases $\lam_{t}$. There is a
range of learning rates over which this dynamic leads to quasi-stable
oscillations of $\lam_{t}$ around the edge of stability value $2/\lr$
(\cref{fig:armijo_theory}, blue curves).

When using a learning rate tuner, $\lr_{t}$ is also a dynamical variable. This
introduces the additional complication of a shifting edge of stability.
Therefore, it is advantageous to analyze the dynamical system using normalized
variables \citep{agarwala_secondorder_2022}. We define $\y_{t}\coloneqq
\lr_{t}\lam_{t}-2$, where $\y = 0$ corresponds to the EOS, and $\p_{t} \coloneqq
\x_{t}^{2}$. This gives us the dynamical equations (\cref{app:theory_deriv})
\begin{equation}
\p_{t+1} = (1+\y_{t})^{2}\p_{t},~\y_{t+1} = \lr_{t+1}\left[\lr_{t}\left(\a-\b\p_{t}\right)\right]+\left(\frac{\lr_{t+1}}{\lr_{t}}\right)\y_{t}+2\left[\frac{\lr_{t+1}}{\lr_{t}}-1\right].
\label{eq:p_y_theory}
\end{equation}
We must then supply a rule for $\lr_{t+1}$. 
In \cref{fig:main_sharp_failures}, we observed that in the full batch setting,
the learning rate multiplied by the sharpness appears to quickly approach a
threshold of $2 - \eps$ (corresponding to $y = -\eps$), and then varies slowly
below the EOS threshold. We model the varying learning rate as
\begin{equation}
\lr_{t} \coloneqq 2(1-\eps)/\lam_{t}\,.
\label{eq:armijo_approx}
\end{equation}
This maintains $\y_{t}=-\eps$. Notably, this schedule was explicitly proposed
by \citet{cohen2021gradient} (see also \cref{fig:exact_edge}). In this regime,
$\p_{t}$ decreases monotonically, aligning with the original goal of these
methods to decrease the loss (\cref{fig:p_dynamics_armijo}). However, this
eliminates feedback for controlling the increase in $\lam_{t}$, resulting in
significant progressive sharpening (\cref{fig:armijo_theory}, orange curve).

Consequently, when attempting to enforce monotonicity, learning rate tuners may
inadvertently disrupt the non-linear stabilization that makes gradient descent
robust and effective for training deep neural networks. Continually
undershooting the EOS triggers a snowball effect of decreasing learning rate and
increasing sharpness. If there is no corresponding increase in gradient norms,
this causes optimization to slow down.

The poor performance of the classical learning rate tuners
in~\cref{fig:failures_baselines} therefore appear strongly correlated with their
tendency to \emph{undershoot} the edge of stability in the normalized sharpness
coordinate $\y$. In the following, we focus on
understanding tuners that prioritize training at or near the edge of stability.

\section{Optimizing on the Edge of Stability}\label{sec:on_edge}

\begin{figure}[t]
    \centering
    \includegraphics[width=0.48\linewidth]{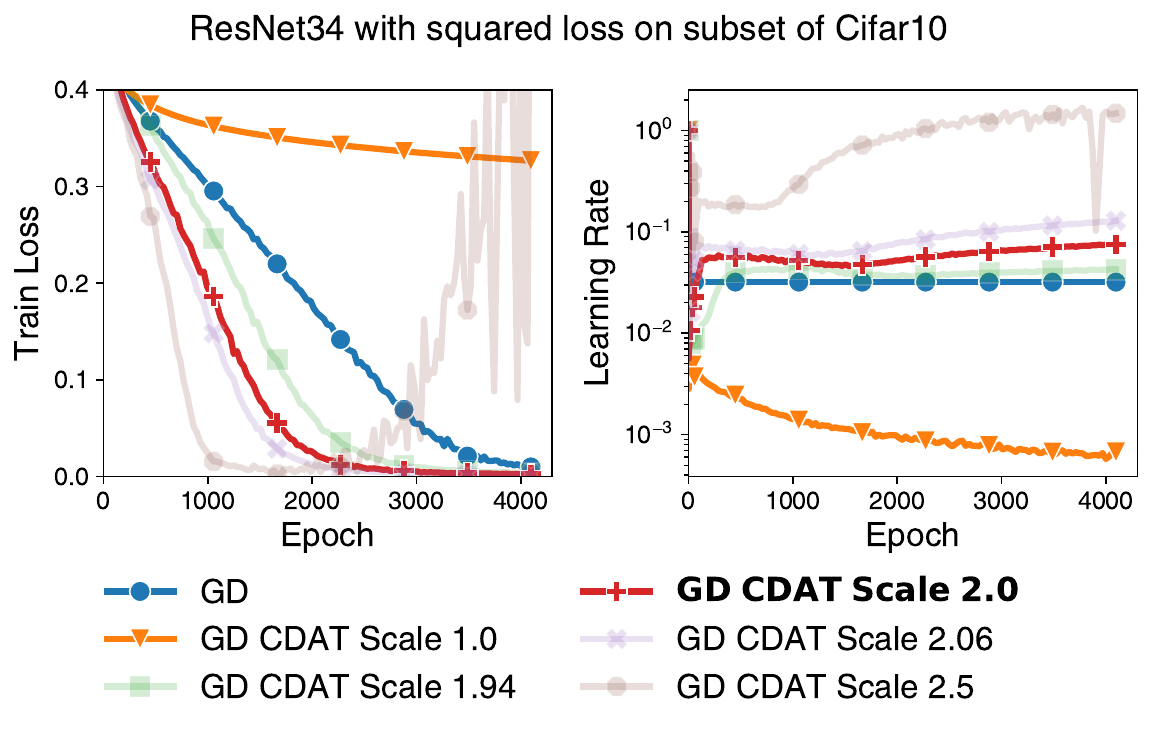}
    \includegraphics[width=0.48\linewidth]{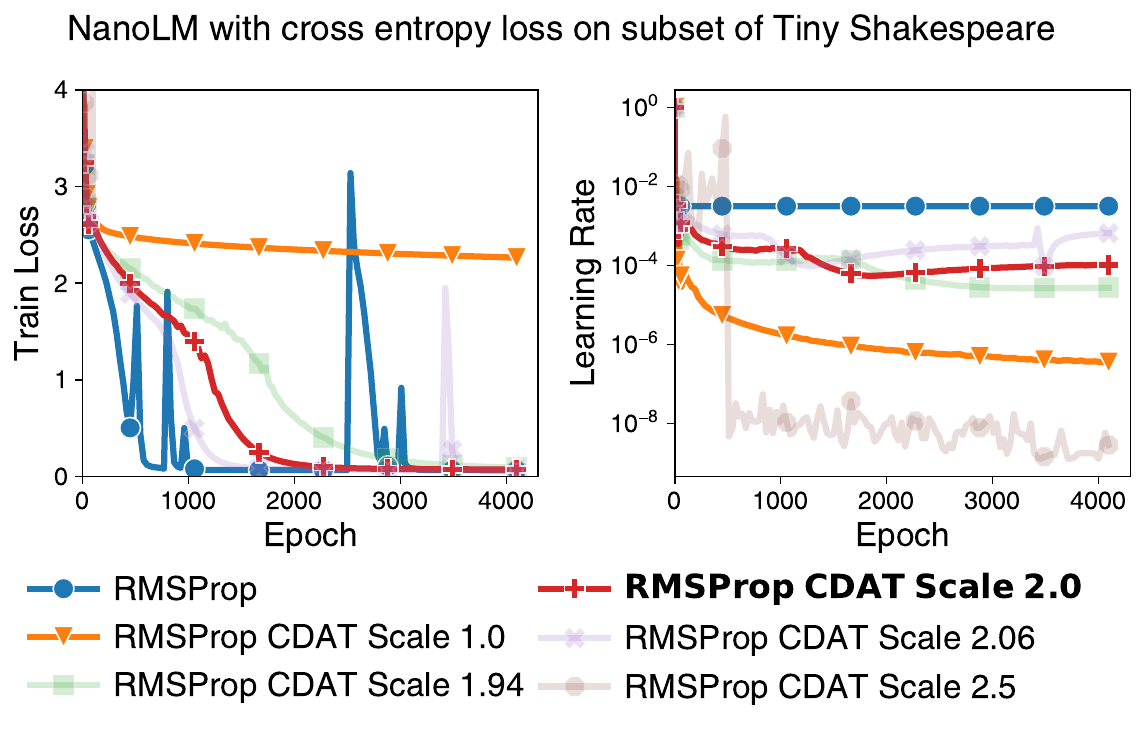}
    \includegraphics[width=0.48\linewidth]{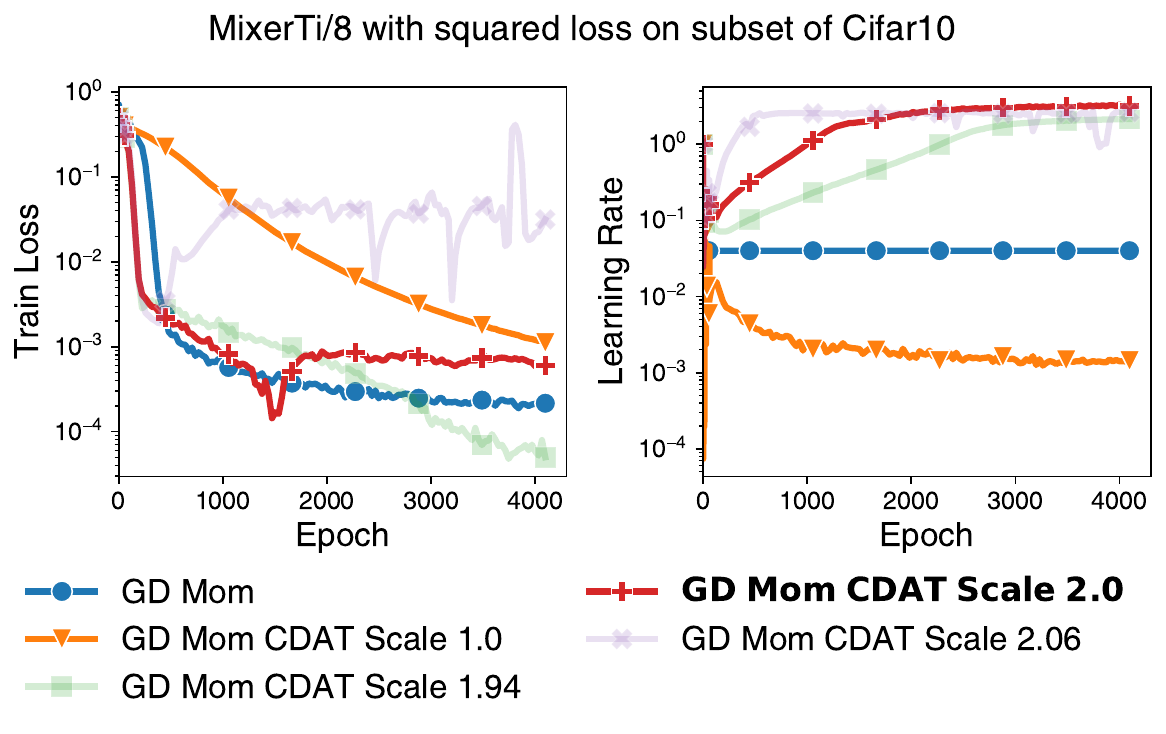}
    \includegraphics[width=0.48\linewidth]{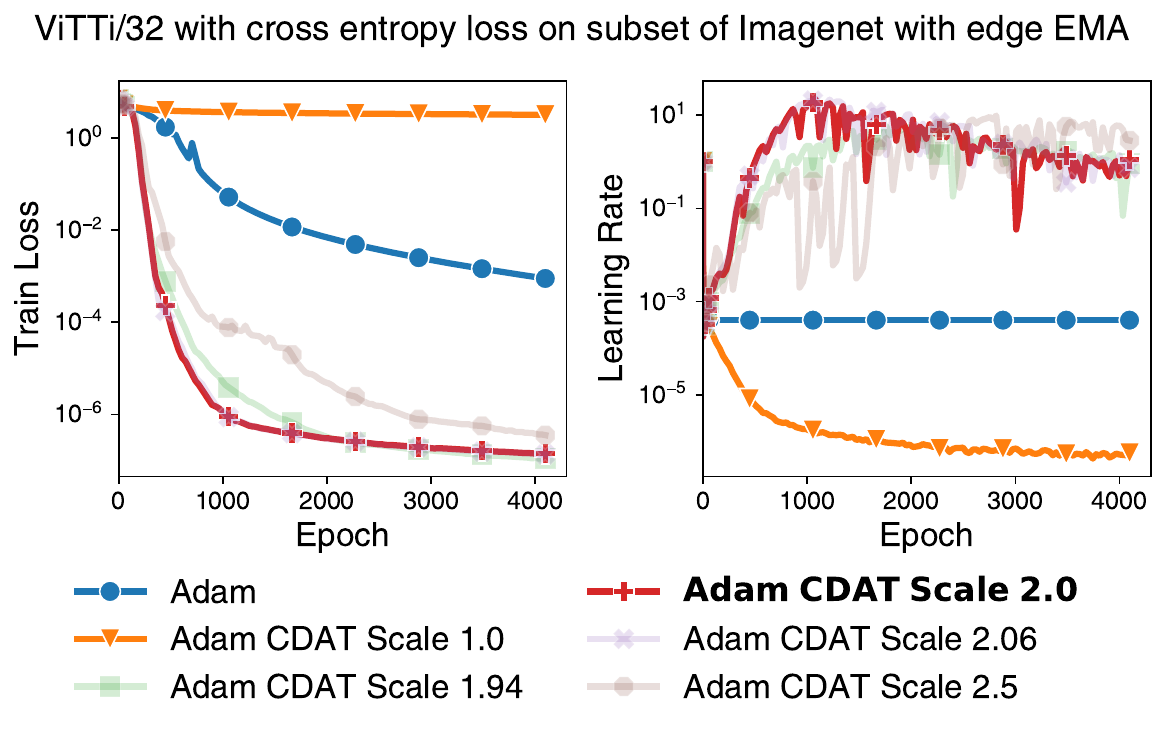}
    \caption{\textbf{Enforcing optimizers to stay on edge} ($\scale=2.0$)
    \textbf{improves performance over greedy approximation} ($\scale=1.0$).
    Train loss and learning rate behaviors for fine-tuned optimizers vs
    self-tuned counterparts with CDAT on various datasets, architectures, losses
    in a full batch regime. Tuning the learning rate ``on edge'' ($\scale\approx
    2$) improves performance over greedy tuning ($\scale= 1$) as well as
    constant learning rate.}
    \label{fig:on_edge}
    \vspace*{-15pt}
\end{figure}

Based on our observations in \cref{sec:classic_tuners}, we design learning rate
tuners that position the underlying optimizer \textbf{on the edge} of stability
($\y= 0$). We analyze a tuner capable of operating both slightly below and
slightly above the EOS in order to exploit nonlinear stabilization.

Formally, we investigate a generalization of the quadratically greedy
rule from \cref{sec:classic_tuners}, which sought $\lr_t$ to minimize the
quadratic approximation $q_f$ in~\eqref{eq:quad_approx}. We instead choose the
learning rate to be \emph{on edge} by seeking the largest value of $\lr$ such
that $q_f$ is smaller or equal to the original value of $f$,
\begin{align}\label{eq:simple_on_edge_rule}
\lr_t^{\text{oe}} & 
\coloneqq 
\max\{\lr \geq 0: 
q_f(\lr; \param_t, u_t) \leq f(\param_t)
\} 
= -2 \frac{\nabla f(\param_t)^\top u_t}{u_t^\top \nabla^2 f(\param_t) u_t}\,,
\end{align}
where the last formula holds provided that $u_t^\top \nabla^2 f(\param_t) u_t>0$
(convex quadratic) and $\nabla f(\param_t)^\top u_t < 0$ ($u_t$ is a descent
direction). 
For $u_t=-\nabla f(\param_t)$, and if $-\nabla f(w_t)$ is aligned
with the eigendirection $v_{\max}$ associated with the largest eigenvalue
$\lam_{\max}$ of $\hmat$, we recover the familiar $\lr_t^{\text{oe}} =
2/\lam_{\max}$. Note however, that contrarily to using directly $\eta_t =
2/\lam_{\max}$, the on-edge rule can naturally take into account the alignment
with $v_{\max}$ (see~\cref{fig:exact_edge}).

We note that the only difference between this and the quadratically greedy rule
is a factor of $2$ in the numerator. Inspired by this observation, and with an
eye towards robustness, we define our \emph{\cdat} (\CDAT) rule by:
\begin{equation}\label{eq:gen_on_edge_rule}
    \lr_t^{\text{cdat}} 
= \scale\frac{n_t}{d_t}, \quad 
\mbox{for} \ n_t = \max\{-\nabla f(\param_t)^\top u_t, 0\}, 
\ d_t = |u_t^\top \nabla^2 f(\param_t) u_t| + \varepsilon.
\end{equation}
The scaling factor $\scale$ lets us interpolate between greedy ($\scale = 1$)
and on-edge ($\scale = 2$). We are most interested in the behavior near $\scale
= 2$. In \eqref{eq:gen_on_edge_rule}, the $\max$ function takes care of the case
where $u_{t}$ is an ascent direction ($\nabla f(\param_t)^\top u_t >0$), the
absolute value takes care of cases where the objective has negative curvature in
the update directions (see~\cref{app:exp_details} for additional justification),
and we simply set $\varepsilon=0$ as we always observed non-negligible positive
curvature. The definitions of the numerator $n_t$ and the denominator $d_t$
allow for the possibility of exponential moving averages (EMA) of each quantity
such as $\tilde n_{t+1} = (1-\beta_{\text{cdat}}) n_t + \beta_{\text{cdat}}
\tilde n_t$ for $\beta_{\text{cdat}}$ referred to as the \CDAT\ EMA parameter
thereafter. We observed that smoothing the estimates of $n_t$ and $d_t$ by an
EMA is particularly relevant when the updates are themselves defined through an
exponential moving average as in Adam, or when using the proposed rule in a
stochastic setting.

\begin{figure}[t]
    \centering
    \includegraphics[width=0.75\linewidth]{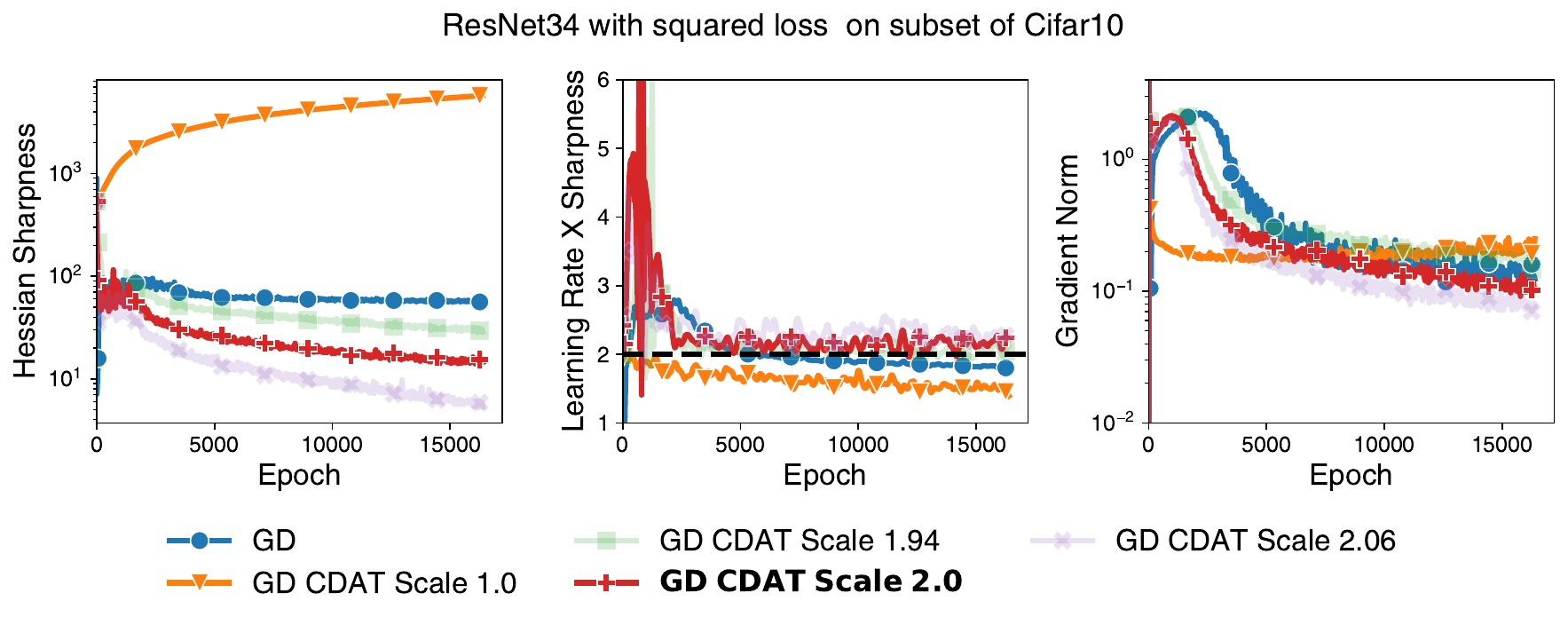}
    \caption{\textbf{Optimizing on edge induces different curvature dynamics.} 
    Sharpness, product between learning rate and sharpness, and gradient norm
    evolutions for gradient descent with CDAT. By putting the learning rate on
    edge ($\scale\approx 2$), the sharpness does not ever increase and actually
    decreases slightly over time. GD with CDAT operates slightly above the edge
    constantly during training. Its gradient norm evolution is akin to a
    fine-tuned constant learning rate baseline. }
    \label{fig:on_edge_sharp}
    \vspace*{-15pt}
\end{figure}

\CDAT{} has two major advantages: it is sensitive to information from all
eigenvalues of $\nabla^2 f(\param_t)$, and it depends on updates $u_{t}$ coming
from any base optimizer. We will take advantage of these properties to explore
the behavior of ``on edge'' optimization in a variety of settings.

\subsection{On edge optimizers in practice}

\paragraph{Full batch regime}

\cref{fig:on_edge} presents results for training with \CDAT{} across various
optimizers, architectures, datasets, and losses. Overall, selecting the learning
rate to be on edge ($\scale=2$) is on par with or better than a fine-tuned
constant learning rate and is always better than a quadratically greedy approach
($\scale=1$). This observation holds even though the quadratically greedy rule
ensures larger instantaneous decrease (\cref{fig:on_edge_vs_greedy}). One notes
that targeting slightly above the edge ($\scale=2.0625$) provides even better
performance than the on edge rule ($\scale=2$) on all examples except the MLP
Mixer on CIFAR10. However, targeting higher above  the edge ($\scale=2.5$)
generally gives diverging results in the short or long terms. Remarkably, all
choices around the edge ($1.9375, 2.0, 20625$) show a progressive increase of
the learning rate that results generally in a better performance than the
constant learning rate counterparts, except for RMSProp on the NanoLM
experiment. The increasing learning rate behavior is akin to the warm-up phase
generally hard-coded by a scheduler. To integrate the proposed rule with the
Adam optimizer, we observed that the estimation of the curvatures through $n_t$,
$d_t$ in~\eqref{eq:gen_on_edge_rule} was necessary.
In~\cref{fig:on_edge_vs_schedules}, we observe that the CDAT rule displays
similar behavior as warm-up schedules, yet it may not fully capture the benefits
of prefixed schedules.

In~\cref{fig:on_edge_sharp}, we analyze the dynamics of the curvature when
optimizing on edge. We observe that the sharpness can be pushed to reduce over
the iterations ($1$\textsuperscript{st} panel of \cref{fig:on_edge_sharp}). The
CDAT rule may operate constantly slightly above the edge
($2$\textsuperscript{nd} panel of \cref{fig:on_edge_sharp}). By reducing the
sharpness, the algorithm may be able to take larger stepsizes and converge
faster. Sensitivity to architecure's width and depth, as well as weight decay,
are also analyzed in~\cref{fig:on_edge_mlp_analysis}.

\begin{figure}[t]
  \begin{minipage}{\linewidth}
    \centering
    \includegraphics[width=\linewidth]{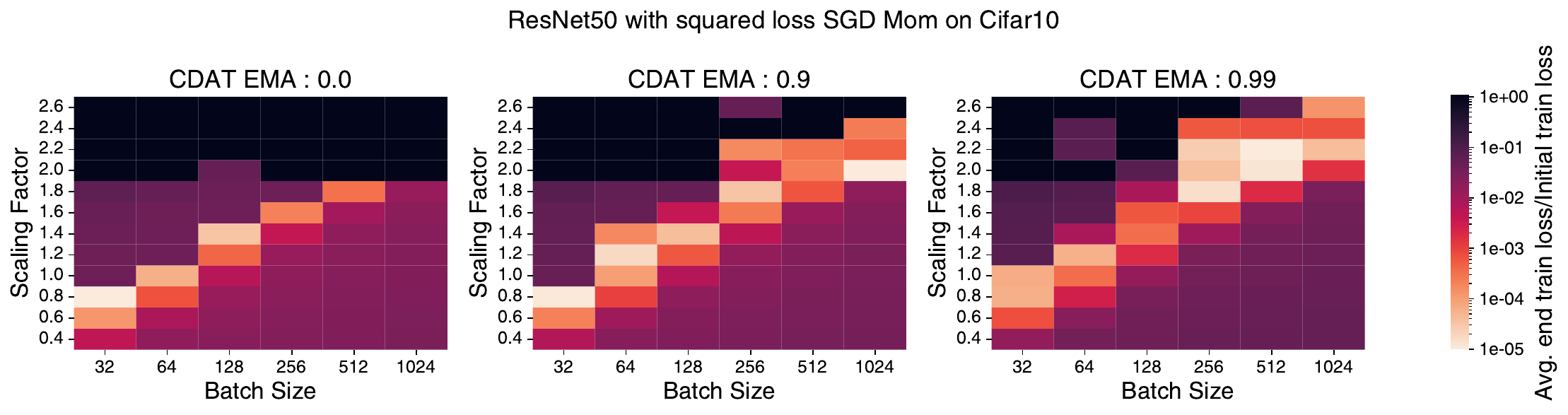}
    \caption{
    \textbf{Stochasticity shifts the optimal scaling.}
    Normalized performance of gradient descent with momentum equipped with CDAT
    in a stochastic regime with varying batch sizes. In a mini-batch regime, the
    optimal scale decreases as the batch size decreases. Using an exponential
    moving average smooths out the performance of the CDAT rule over batch
    sizes.} 
    \label{fig:scaling_factor_vs_batch_size}
  \end{minipage}
  \begin{minipage}{\linewidth}
    \centering
    \vspace*{3pt}
    \includegraphics[width=0.75\linewidth]{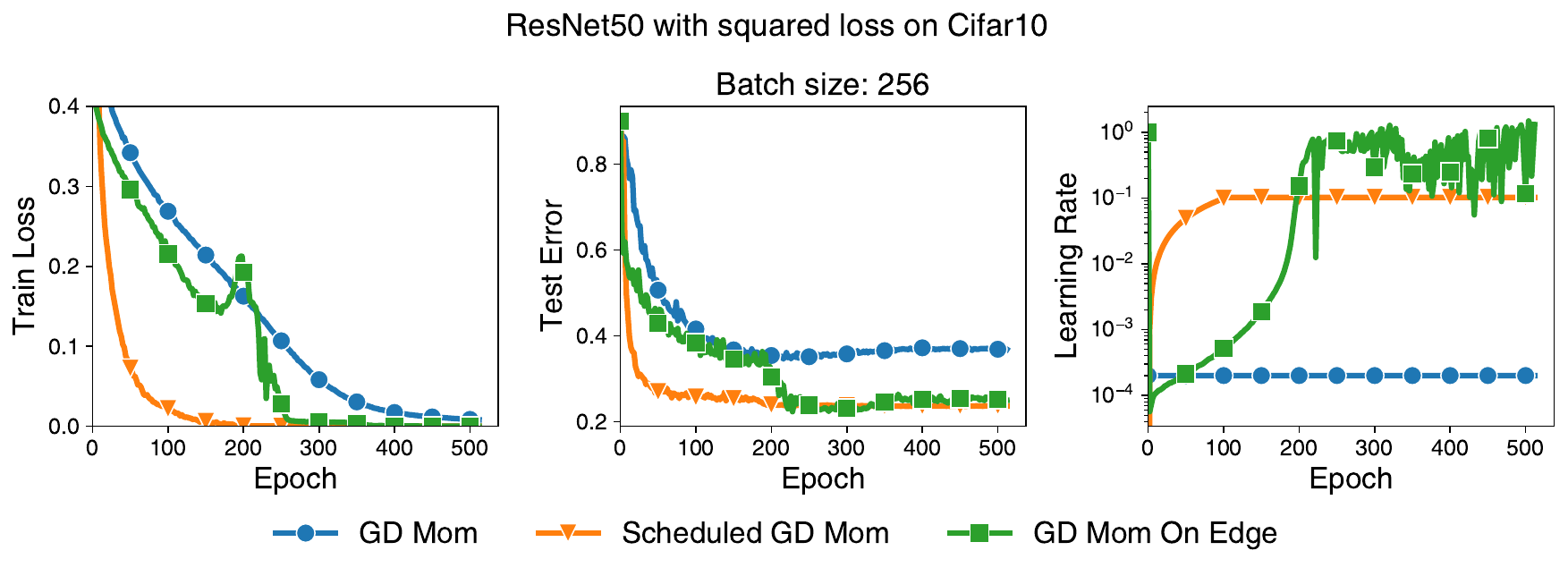}
    \includegraphics[width=0.75\linewidth]{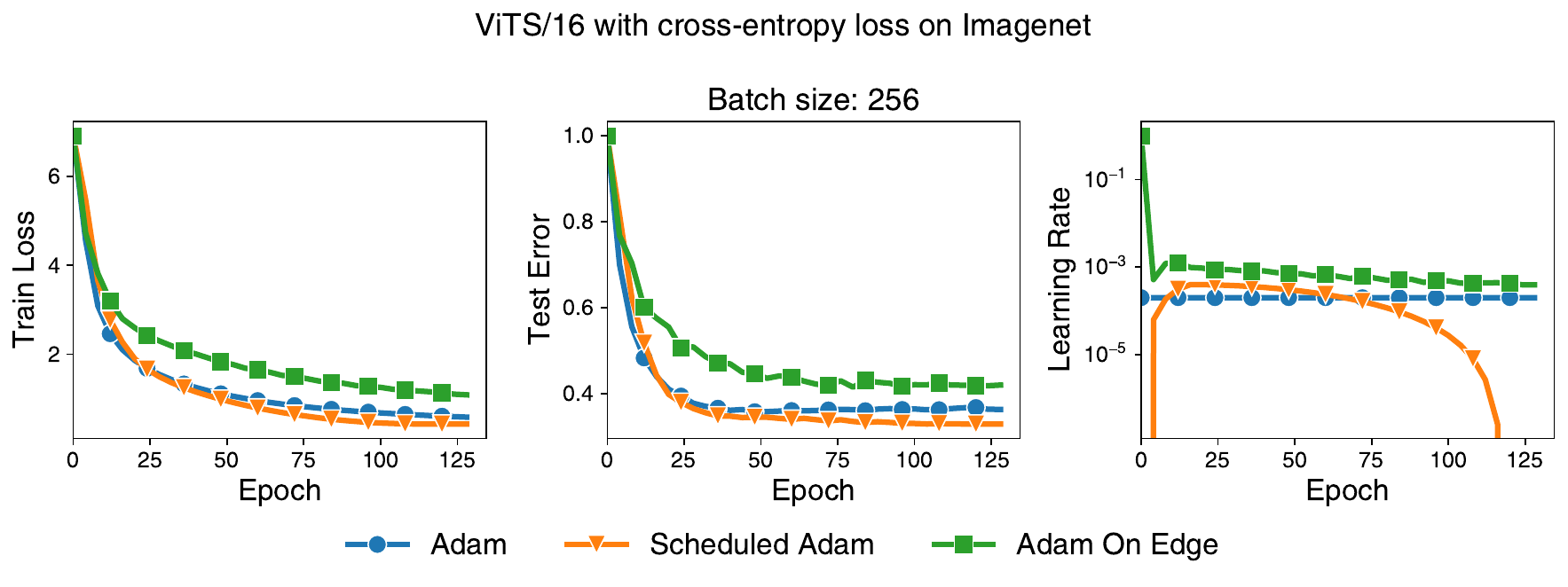}
    \caption{\textbf{The performance of CDAT is subdued in the stochastic regime.}
    Fine-tuned constant, scheduled, and self-tuned with CDAT learning rates in a
    stochastic regime. In a stochastic regime, CDAT can also exhibit a form of
    learning rate warm-up (top figure). However, the interplay between
    sharpening and learning rate are known to be mitigated in a stochastic
    regime which may explain the underperformance of CDAT in this regime (bottom
    figure).}
    \vspace*{-15pt}
    \label{fig:on_edge_stoch}
  \end{minipage}
\end{figure}

\paragraph{Mini batch regime}
The CDAT rule can be used in a stochastic regime by replacing $f$
in~\eqref{eq:gen_on_edge_rule} by its stochastic counterpart $f^{(m_t)}$ on a
mini-batch $m_t$. However, two difficulties may arise.

First, the on edge rule is motivated by the sharpening effects of the overall
objective, which can be overestimated or underestimated by a single mini-batch.
As a result the optimal scaling factor may vary with the mini-batch. In
\cref{fig:scaling_factor_vs_batch_size}, we observe that the optimal scaling of
the on-edge rule is proportional to the batch size up to some size. In
particular, at specific batch sizes, we observe that the greedy rule
($\scale=1$) outperforms the on-edge rule. This result is consistent with the
good performance of linesearches or greedy rules in a mini-batch regime
previously mentioned and observed in~\cref{fig:baselines_stoch}. We also observe
in~\cref{fig:scaling_factor_vs_batch_size}, that integrating an EMA into the
estimation of the edge in~\eqref{eq:gen_on_edge_rule} smooth out the selection
of the optimal scaling factor.

Second, the sharpening effects are known to be generally mitigated in the
stochastic regime \citep{jastrzebski_three_2018, cohen2021gradient,
agarwala_high_2024}. The benefits of the on edge rule appear also subdued in
this regime
(\cref{fig:on_edge_stoch},~\cref{fig:resnet50_stoch_all_bs},~\cref{fig:vit_adam_stoch_all_bs}).

\subsection{Modeling \CDAT{} dynamics}

\label{sec:on_edge_theory}

\begin{figure}[t]
  \centering
  \begin{tabular}{ccc}
  \includegraphics[height=0.25\linewidth]{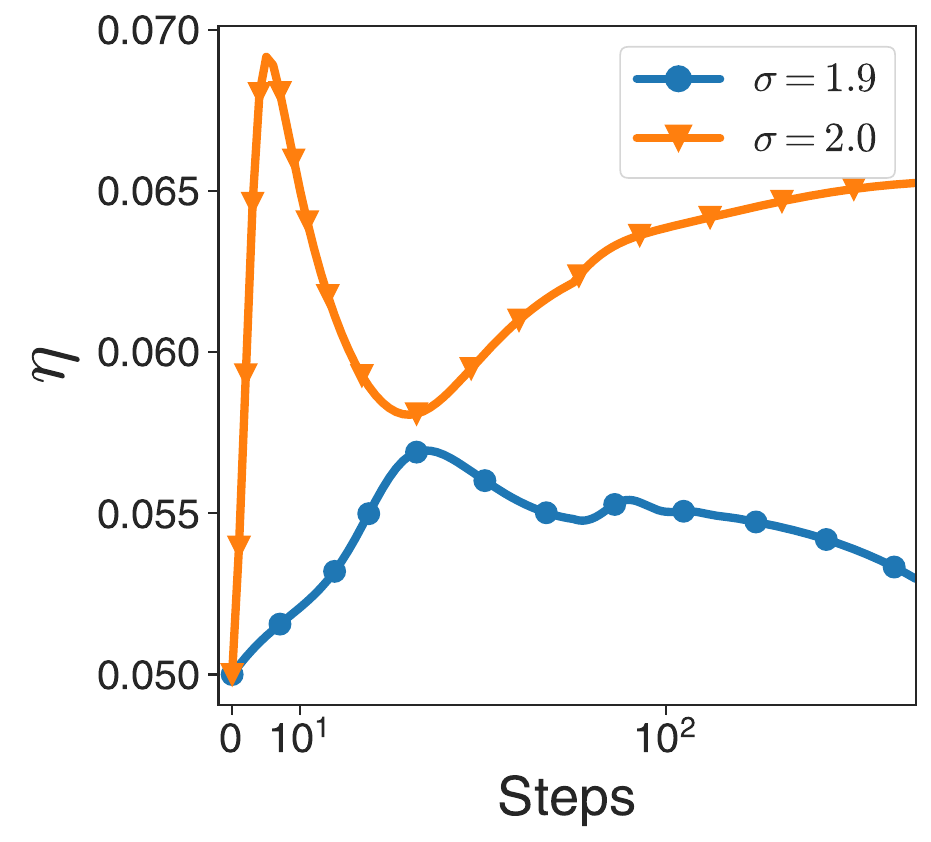} &
  \includegraphics[height=0.25\linewidth]{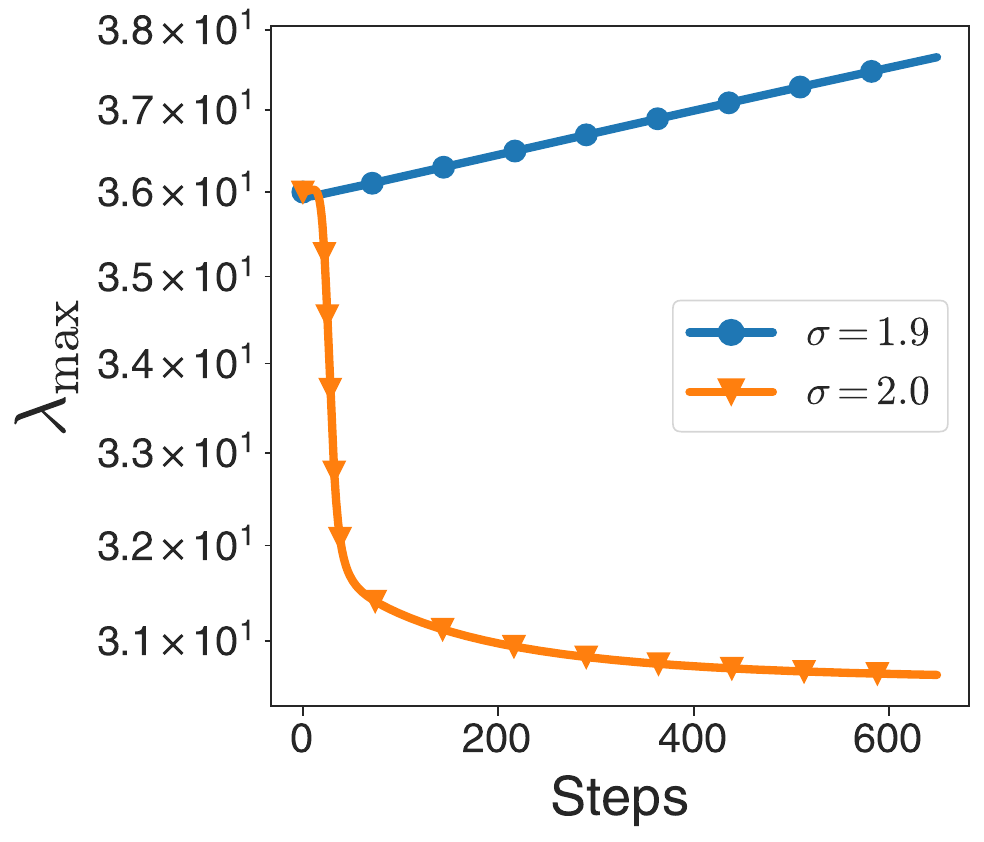} &
  \includegraphics[height=0.25\linewidth]{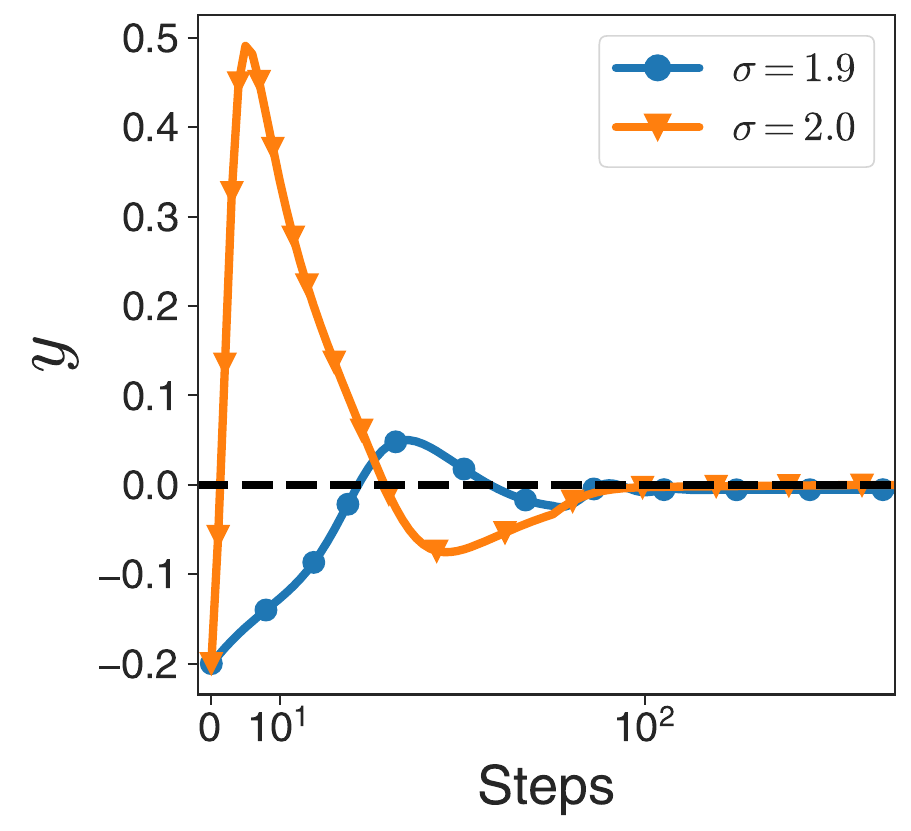}
  \end{tabular}
  \caption{
  \textbf{A simple model partially captures the benefits induced by the proposed CDAT rule.}
  Dynamics of theoretical model of \CDAT{}~\eqref{eq:cdat_model}. For $\scale =
  2$, feedback stabilizes $\y$ close to the EOS ($\y = 0$), which stabilizes
  $\lam_{\max}$ (orange). For $\scale = 2-\eps$ and small $\eps$ (blue, $\eps =
  0.1$), model predicts that $\lam_{\max}$ slowly grows (middle), but predicts
  that $\y$ stabilizes to a value $-\eps\ll\y_{t}<0$ (right).
  }
  \vspace*{-15pt}
  \label{fig:edge_theory}
\end{figure}

The classical optimization framework is insufficient to fully explain the
benefits of \CDAT{}. For example, on a convex quadratic objective, $\scale = 1$
is the optimal choice, and $\scale>2$ results (in the worst case) in a divergent
algorithm. However, we can use a simplified model to begin understanding the
joint dynamics of the learning rate and sharpness under \CDAT{}.

We approximate the gradients around a stationary point $\param_\star$, where
$\nabla f(w_\star) = 0$, as $\nabla f(w_t) \approx H \bar \param_{t}$ for $\bar
\param_t \coloneqq \param_t - \param_\star$, and $H$ being a symmetric matrix.
In this scenario, the learning rate  given by \CDAT{} is $\lr_t^{\text{cdat}} =
\scale(\bar \param_t^{\top}\hmat^{2}\bar \param_t) / (\bar
\param_t^{\top}\hmat^{3}\bar \param_t)$. Consider the case where $\hmat$ has two
eigenvalues $\lam$ and $\nu$, with $\lam>\nu \geq 0$. In this case the \CDAT{}
learning rate can be written as
\begin{equation}
\lr_t^{\text{cdat}} = \scale\frac{\lam^{2}\p_{t}+\nu^{2}\g_{t}}{\lam^{3}\p_{t}+\nu^{3}\g_{t}} = \scale\frac{\lam^{2}\p_{t}/\g_{t}+\nu^{2}}{\lam^{3}\p_{t}/\g_{t}+\nu^{3}}\,.
\label{eq:oe_stepsize_theory}
\end{equation}
Here $\p_{t}, g_t$ are the projections $p_t := (\bar \param_t^\top \v)^{2},
\g_{t} \coloneqq (\bar \param_t^\top \w)^{2}$, respectively onto the
eigendirections $\v$ and $\w$ associated with $\lambda$, $\nu$. Therefore,
$\lr_t^{\text{cdat}}$ interpolates between its minimum value $\scale/\lam$ to
the larger value $\scale/\nu$, depending on the alignment ratio $\p_{t}/\g_{t}$.
For $2\nu/\lam<\scale<2$, this rule can achieve learning rates both above and
below the EOS.

We can gain additional insight by modeling a dynamical $\lam_{t}$, extending the
model of \cref{sec:theory_part_1}. While model~\eqref{eq:p_y_theory} captures
the dynamics in the largest eigendirection $\v$, here we aim to model the
dynamics in the orthogonal subspace. To simplify, we consider the
eigendirections $\v, \w $, and small eigenvalue $\nu$ fixed. We then model the
gradients as  $\nabla f(w_t) \approx H_t \bar \param_{t}$ with $H_{t} =
\lam_{t}\v\v^{\tpose}+\nu \w\w^{\tpose}$. If we update $\param$ in the direction
$\v_{\perp}$ using gradient descent on $\nu\g_{t}$, we obtain the following
dynamical system describing the \CDAT{} learning rate tuner:
\begin{equation}\label{eq:cdat_model}
\eta_{t+1} =  \scale\frac{\lam_{t}^{2}\p_{t}+\nu^{2}\g_{t}}{\lam_{t}^{3}\p_{t}+\nu^{3}\g_{t}}, \quad
\g_{t+1} = (1-\lr_{t}\nu_{t})^{2}\g_{t}, \quad
\p_{t+1} = (1+\y_{t})^{2}\p_{t}\,.
\end{equation}
Combining this with the update rule for $y_t$ given in \eqref{eq:p_y_theory}
completes the model.

There are two important regimes of behavior in this model. First, if $\y_{t}>0$,
$\p_{t}$ will increase and eventually $\y_{t}$ will decrease as in the normal
EOS case. If $\y_{t} < 0$, the key threshold is $\y_{t}< - \lr_{t}\nu_{t}$. In
this case, the ratio $\p_{t}/\g_{t}$ \emph{decreases} - leading to an increase
in $\lr_{t}$ according to the on edge rule. If $a-b\p_{t}>0$ (as it is if
$\p_{t}$ has become small due to $\y_{t}<0$), then we see from
\eqref{eq:p_y_theory} that this leads to an \emph{increase} in $\y_{t}$. This
suggests that \CDAT{} has a tendency to push $\y_{t}$ closer to the EOS --
sending $\y$ towards $0$ if the learning rate is driven by the eigendirections
corresponding to smaller eigenvalues.

Numerical simulations on this model (\cref{fig:edge_theory}) suggest that this
effect can indeed cause remarkably small values of $\y$ ($3$\textsuperscript{rd}
panel of \cref{fig:edge_theory}). We emphasize that this is due to the
\emph{joint dynamics} of $\lr_{t}$ (induced by the learning rate tuner), and
$\lam_{t}$, $\p_{t}$, and $\g_{t}$ (induced by GD). There are also important
limitations in this model's ability to fully explain \CDAT{}'s behavior. For
example, the model predicts runaway sharpening for $\scale<2$
($2$\textsuperscript{nd} panel of \cref{fig:edge_theory}), and divergence for
$\scale>2$. In practice, we saw a range of stable and useful settings for scale
centered around $2$. This modeling limitation likely stems from neglecting the
dynamics orthogonal to $\v$ as well as higher-order terms, which empirically
tend to stabilize EOS dynamics \citep{agarwala_secondorder_2022}. 

\section{Conclusion and Future Directions}
\label{sec:dyn_theory_discuss}

\paragraph{Summary}
Our empirical results showed that simple linesearches and approximate greedy
learning rate tuners underperform constant learning rate approaches in the full
batch regime -- despite being better on individual steps. The idea that
``locally greedy'' methods perform poorly on long time scales has been shown in
other settings as well, including evolutionary dynamics
\cite{agarwala_adaptive_2019}. Our experiments and theoretical work suggest the
failure of these classical tuners is due to the fact that they suppress the
feedback which stabilizes sharpness in the fixed learning rate setting. As the
sharpness increases, tuners are forced to take smaller steps, which ends up
leading to slower learning.

We find, in contrast, that prioritizing stability of the sharpness yields
tangible benefits. Our CDAT method pushes the network towards the edge of
stability via a dynamically driven process. It also naturally displays some form
of progressive increase of the learning rate akin to prefixed warm-up schedules.
CDAT also sheds light on the more complicated dynamics in small mini batch
regime, where estimation of a locally greedy rule may actually place the
optimizer on the edge of stability of the full batch objective.

\paragraph{Limitations and future directions}
We explored some limitations of the current modeling framework in Section
\ref{sec:theory_part_1} -- in particular, the failure to capture stabilization
due to higher order terms. Developing improved models (either analytically or
numerically) would allow for powerful tools from other disciplines to aid
algorithm design -- particularly, methods from control theory. For example,
state feedback schemes can be designed through the analysis of nonlinear
dynamical systems to ensure asymptotic stabilization~\citep[Chapter
7]{isidori1985nonlinear}. We believe a cross disciplinary approach will be
useful for designing the next generation of learning rate tuners.

\newpage
\bibliography{stepsize_tuning_refs}
\bibliographystyle{unsrtnat}

\newpage
\appendix

\newpage

\section{Theoretical Model}

\subsection{Dynamics of rescaled variables}

\label{app:theory_deriv}

In Section \ref{sec:theory_part_1}, we used Equation \ref{eq:x_l_theory} to
derive equations in $\p\coloneqq \x^{2}$ and $\y\coloneqq \lr\lam-2$. We arrived
at
\begin{equation}
\p_{t+1} = (1+\y_{t})^{2}\p_{t},~\y_{t+1} = \lr_{t+1}\left[\lr_{t}\left(\a-\b\p_{t}\right)\right]+\left(\frac{\lr_{t+1}}{\lr_{t}}\right)\y_{t}+2\left[\frac{\lr_{t+1}}{\lr_{t}}-1\right].
\end{equation}

The dynamical equation for $\p$ is obtained by squaring the equation for
$\x_{t}$ in Equation \ref{eq:x_l_theory}. To derive the equation for $\y_{t}$,
we first derive the dynamics of $\lr_{t}\lam_{t+1}$ as
\begin{equation}
\lr_{t}\lam_{t+1} = \lr_{t}\left[\lr_{t}\left(\a-\b\p_{t}\right)\right]+\lr_{t}\lam_{t}.
\end{equation}
We then have
\begin{equation}
\y_{t+1} = \frac{\lr_{t+1}}{\lr_{t}}[\lr_{t}\lam_{t+1}-2]+2.
\end{equation}
Evaluating completes Equation \ref{eq:p_y_theory}.

\subsection{Dynamics of the projection on the largest eigenvector}

We present results on the dynamics of $\p$ in the various models; these were
omitted from the main text in order to simplify the presentation. For constant
learning rate, $\p$ initially decreases until EOS is crossed, after which it
enters into a cycle of increase and decrease (Figure
\ref{fig:p_dynamics_armijo}, blue). For our model of linesearches, where
$\lr_{t} = 2(1-\eps)/\lam_{\max, t}$, $\p$ decays to $0$ quickly and there is no
mediation of sharpening (orange, $\eps = 0.1$).

For our model of \CDAT{} presented in Section \ref{sec:on_edge_theory}, $\p$
stabilizes for $\scale = 2$ (Figure \ref{fig:p_dynamics_on_edge}, orange). For
$\scale<2$, the model still predicts decay of $\p$, but the ratio of $\p_{t}$ to
the orthgonal component $\g_{t}$ remains constant (Figure
\ref{fig:p_dynamics_on_edge}, blue). This fixed ratio stabilizes $\y$ to a value
near $0$.

In practice, the higher order terms in the dynamics provide additional
stability, in the on-edge model, which allows $\p$ to stabilize as well,
see~\cref{fig:app_sharp_failures},~\cref{fig:on_edge_more_metrics}. The key is
that these terms can operate when $\y$ is close to $0$ for long periods of time.
These results suggest that additional model development is required to
understand the behavior of learning rate tuners which target the EOS.

\section{Additional Experiments}\label{app:more_exp}

\subsection{Further analyzes of base learning rate tuners}
\cref{fig:app_sharp_failures} completes \cref{fig:main_sharp_failures} with
measures of sharpening and learning rates on the settings considered
in~\cref{fig:failures_baselines}. For RMSProp we considered the preconditioned
Hessian following the observations done by~\citet{cohen2022adaptive} that for
adaptive gradient methods such as RMSProp or Adam, the sharpness of the
preconditioned Hessian, rather than the sharpness of the Hessian, defines the
edge of stability. Namely, recall that RMSProp takes updates of the form 
\[
w_{t+1} = w_t - P_t^{-1} \nabla f(w_t), \quad \mbox{for} P_t = \diag(\sqrt{\nu_t + \varepsilon})
\]
for 
\[
\nu_t = (1-\beta_2)g_t^2 + \beta_2 \nu_{t-1}, \quad \nu_{-1} = 0, \ g_t = \nabla f(w_t),
\]
with $\beta_2$ an exponential moving average parameter. The preconditioned
Hessian takes then the form
\[
 \tilde H_t = P_t^{-1/2} \nabla^2 f(w_t) P_t^{-1/2},
\]
and we report $\lambda_{\max}(\tilde H_t)$.

We observe similar behaviors in these regimes as
in~\cref{fig:main_sharp_failures}. Namely, the sharpness or preconditioned
sharpness ever increase ($2$\textsuperscript{nd} panels of
\cref{fig:app_sharp_failures}), while the learning rates ever decrease
($1$\textsuperscript{st} panels of \cref{fig:app_sharp_failures}). The constant
learning counterpart can operate above the edge of stability while the
self-tuned methods generally avoid crossing the edge of stability
($3$\textsuperscript{rd} panels of \cref{fig:app_sharp_failures}).

\begin{figure}[t]
\begin{minipage}{0.48\textwidth}
\centering
\includegraphics[width=0.62\linewidth]{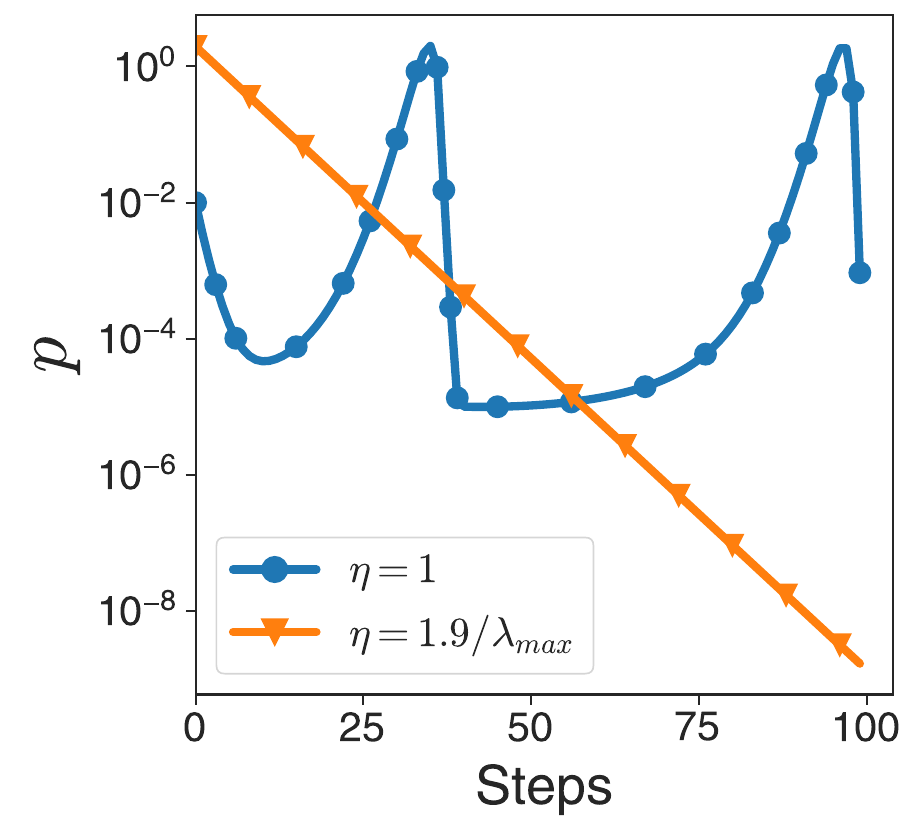}
\caption{Dynamics of the largest eigenvalue projection $\p$. For constant learning
rate, $\p$ cycles between stability and instability and $\lam_{\max}$ stabilizes
(blue). If learning rate tuner sets $\lr_{t} = 2(1-\eps)/\lam_{\max, t}$, $\p$
decays to $0$ and there is no negative feedback preventing sharpening.
\label{fig:p_dynamics_armijo}
} \end{minipage}\hfill 
\begin{minipage}{0.48\textwidth} 
\centering
\includegraphics[width=0.62\linewidth]{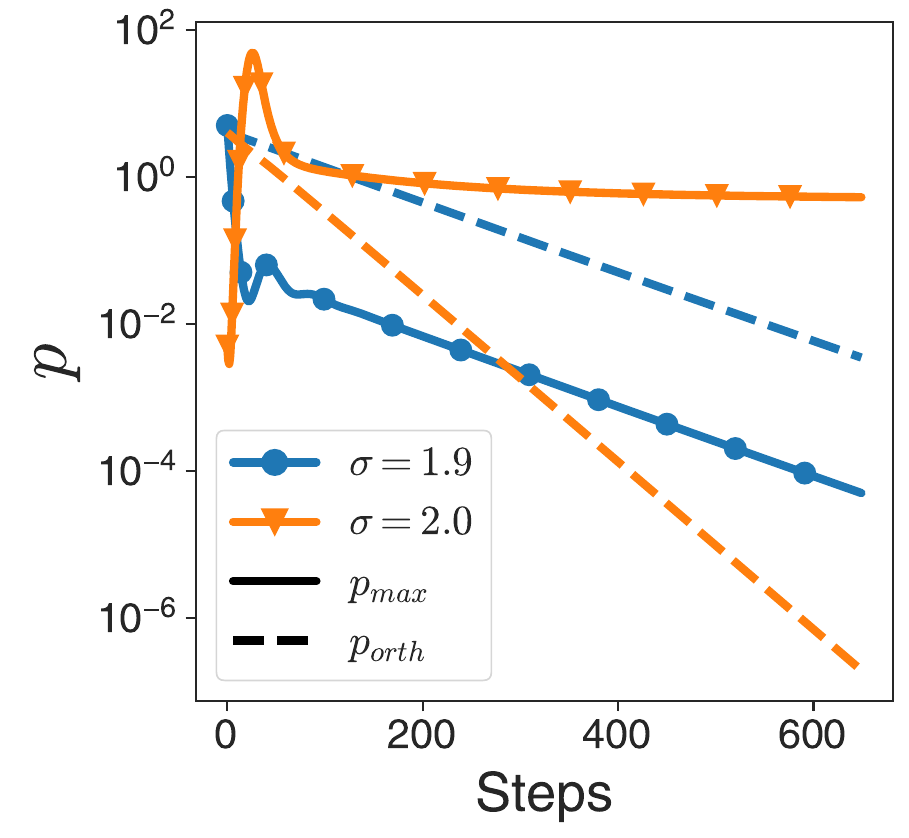}
\caption{ Dynamics of $\p$ for toy model of \CDAT{}. For $\scale = 2$, $\p$ is
stable, which induces stabilization of sharpness (orange). For $\scale = 2-\eps$
for small $\eps$ (blue, $\eps = 0.1$), $\p$ decreases but the ratio of $\p$ to
the orthogonal projection remains constant which stabilizes $\y$ near $0$. 
\label{fig:p_dynamics_on_edge}
}
\end{minipage}
\end{figure}

\subsection{Analyzing additional learning rate tuners}
We consider the performance of two additional classical learning rate tuners,
Polyak stepsize~\citep{polyak1964some, berrada2020training,
loizou2021stochastic} and hyper-gradient descent~\citep{almeida1999parameter,
baydin2017online} akin to the resilient backpropagation
scheme~\citep{riedmiller1992rprop}. 

Briefly, Polyak stepsizes consider setting the learning rate as
\begin{equation}\label{eq:polyak}
    \lr_t = \min\left\{\frac{f(w_t) - f^\star}{\|\nabla f(w_t)\|^2}, \lr_{\max}\right\},
\end{equation}
where $f^\star = \min_w f(w)$ is the minimum of the objective set to $0$ by
assuming that a neural network can overfit the data, and $\lr_{\max}$ is a
maximal stepsize selected as $1$ or $100$ in our experiments (we take the best
instance).

Hyper-gradient descent considers updating the stepsize towards maximal decrease
of the objective. Namely, defining the objective obtained after one step
$h_t(\lr) = f(w_t + \lr u_t)$, the algorithm updates $\lr_t$ by a gradient step
on $h_t$ resulting a priori in $\lr_{t+1} = \lr_t - \alpha \nabla f(w_t + \lr_t
u_t)^\top u_t$ for a given hyper-learning rate $\alpha$.
\citet{almeida1999parameter, baydin2017online} argued for using multiplicative
updates of the form 
\begin{equation}
\label{eq:hypergradient}
\lr_{t+1} 
= \lr_t\left(1- \beta \frac{ \nabla f(w_t + \lr_t u_t)^\top u_t}{\|\nabla f(w_t + \lr_t u_t)\|_2 \|u_t\|_2}\right).
\end{equation}
Intuitively, the learning rate increases if the update is aligned with the
negative gradient direction and decreases otherwise. Resilient
backpropagation~\citep{riedmiller1992rprop} adopts a similar logic
componentwise. In our experiments we vary $\beta$ and select the best instance,
see~\cref{ssec:details_per_figure} for more details.

We observe that Polyak stepsizes (top figure of \cref{fig:other_tuners})
generally select larger learning rates than the constant learning rate
counterpart. The efficiency of Polyak stepsizes is not reached by the CDAT rule
with $\sigma=2$ but with a slightly larger scale $\sigma=2.06$. The efficiency
of the Polyak stepsize method, in particular compared to a simple linesearch, in
a full batch regime, has also been reported by~\citet{roulet2023interplay}. The
proposed CDAT rule may capture the benefits of aggressive learning rates taken
by Polyak stepsizes in a smoother way by allowing various scales.

On the other hand, the hyper-gradient descent performs just on par with the
fine-tuned constant learning rate counterpart (bottom figure
of~\cref{fig:other_tuners}). We also observe a slow, yet steady, progressive
sharpening when using the hyper-gradient descent. As with the linesearch method
or the quadratically greedy rule, the hyper-gradient descent focuses on
selecting a learning rate that decreases the loss, which appears, across those
tuners, to potentially suppress effective stabilization effects naturally
appearing with constant learning rate.

\subsection{Base learning rate tuners in a stochastic regime}
In~\cref{fig:baselines_stoch}, we report the performance of classical learning
rate tuners (linesearch or quadratically greedy method) in a stochastic regime
for varying batch-sizes. As observed previously by \citet{vaswani2019painless}
or~\citet{roulet2023interplay}, a linesearch for example can perform well in a
stochastic regime. Note that the two approaches (linesearch and quadratically
greedy method) display similar behaviors (just as they displayed similar
behaviors in the full batch regime). This hints that, rather than playing with
the numerous hyperparameters of a linesearch we may focus simply on an
additional scaling factor for the quadratically greedy rule, which motivated the
proposed CDAT rule.

\subsection{Targeting the edge of stability using the exact sharpness}
\citet[Appendix F]{cohen2021gradient} reported bad performance of adaptive
learning rate tuners selecting the stepsize as 
\[
    \lr = 2/\lambda_{\max}(\nabla^2 f(w))\,,
\]
which may fix the learning rate just at the edge of stability. Note that such a
definition does not take into account the additional alignment of the update
with the largest eigenvector. Our proposed diagnostic rule CDAT rather considers
the edge of stability given by a local approximation of the objective along the
update so to take into account the alignment of the update with the largest
eigenvector of the Hessian. We ran experiments with a rule 
\begin{equation}\label{eq:on_exact_edge}
    \lr = \scale/\lambda_{\max}(\nabla^2 f(w))\,,
\end{equation}
that lets the scaling factor vary just as done with CDAT. The only difference is
in the estimation of the base estimate of the edge of stability (CDAT does it
with the help of a quadratic approximation of the objective, while the
rule~\eqref{eq:on_exact_edge} uses an exact computation of the sharpness). In
\cref{fig:exact_edge}, we observe that setting the scale $\sigma\approx 2$ leads
to poor performance as previously observed by~\citet{cohen2021gradient}. Note
however that by setting the scaling much above $2$ (like $\sigma=3$) such a rule
may outperform a constant learning rate. This hints that the
rule~\eqref{eq:on_exact_edge} misses the alignment of the update with the
largest eigenvector, which motivated the CDAT rule.

\subsection{Analyzing instantaneous gains versus long-term gains}
In~\cref{fig:on_edge_vs_greedy}, we investigate the difference of instantaneous
decrease using the quadratically greedy rule (CDAT with $\scale=1$) compared to the on
edge rule (CDAT with $\scale=2$). Throughout a training, the quadratically rule
ensures a larger instantaneous decrease as intended through its definition as a
learning rate that minimizes the loss. Yet, in the long term, the quadratically
greedy rule underperforms the on edge rule (\cref{fig:on_edge}).

\subsection{Additional metrics for the CDAT rule}
In~\cref{fig:on_edge_more_metrics}, we additionally measure the alignment of the
updates with the largest eigenvector and the angle between successive updates.
We observe that the CDAT rule for $\scale \approx 2$ behaves similarly as the
constant learning rate counterpart. In particular, the updates tend to quickly
be in opposed directions. The quadratically greedy rule does not demonstrate
such a behavior.

\subsection{Sensibility analysis to architecture hyperparameters} 
In \cref{fig:on_edge_mlp_analysis}, we study CDAT for simple MLPs in a full
batch regime on the MNIST dataset. Our goal is to understand the benefits of the
proposed CDAT rule for varying hyperparameters. First, we analyze the
sensibility to width and depth of an MLP in a similar fashion as~\citet[Appendix
D]{cohen2021gradient} did to analyze progressive sharpening.

We observe that accrued gains can be obtained with the CDAT rule for larger
widths (top left panel of~\cref{fig:on_edge_mlp_analysis}). Note that
\citet[Appendix D]{cohen2021gradient} found less sharpening at higher widths. In
terms of depth (top right panel of~\cref{fig:on_edge_mlp_analysis}), the CDAT
rule works best with larger depths while we note a slight shift of optimal
scaling factors from $2$ to slightly below $2$.

The CDAT rule appears to work best with small or no weight decay (bottom left
panel of~\cref{fig:on_edge_mlp_analysis}) while its benefits fade with larger
weight decay (no difference between greedy $\scale=1$ and on edge $\scale=2$).
Finally, while the method naturally finds smaller train losses with larger
subsets of data, we do not observe a significant shift of relative performance
between scales as the size of the data increases~(bottom right panel
of~\cref{fig:on_edge_mlp_analysis}).

\subsection{CDAT rule versus prefixed schedule in full batch regime}
\begin{figure}[t]
    \centering
    \includegraphics[width=0.48\linewidth]{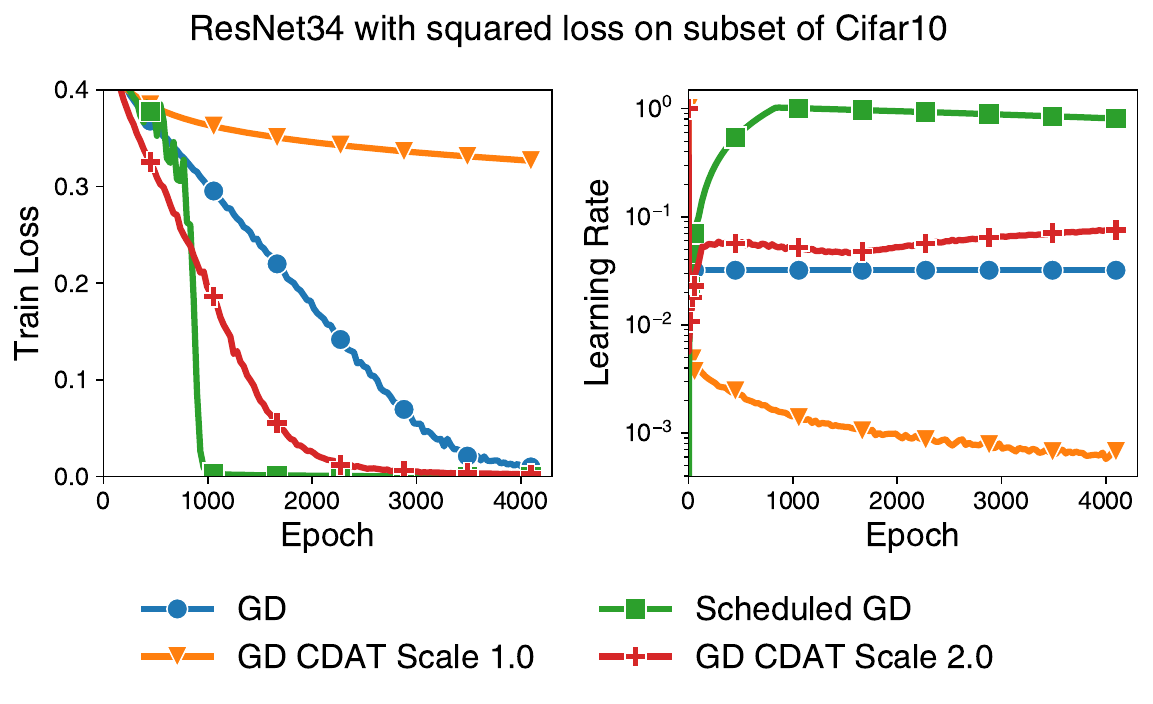}
    \includegraphics[width=0.48\linewidth]{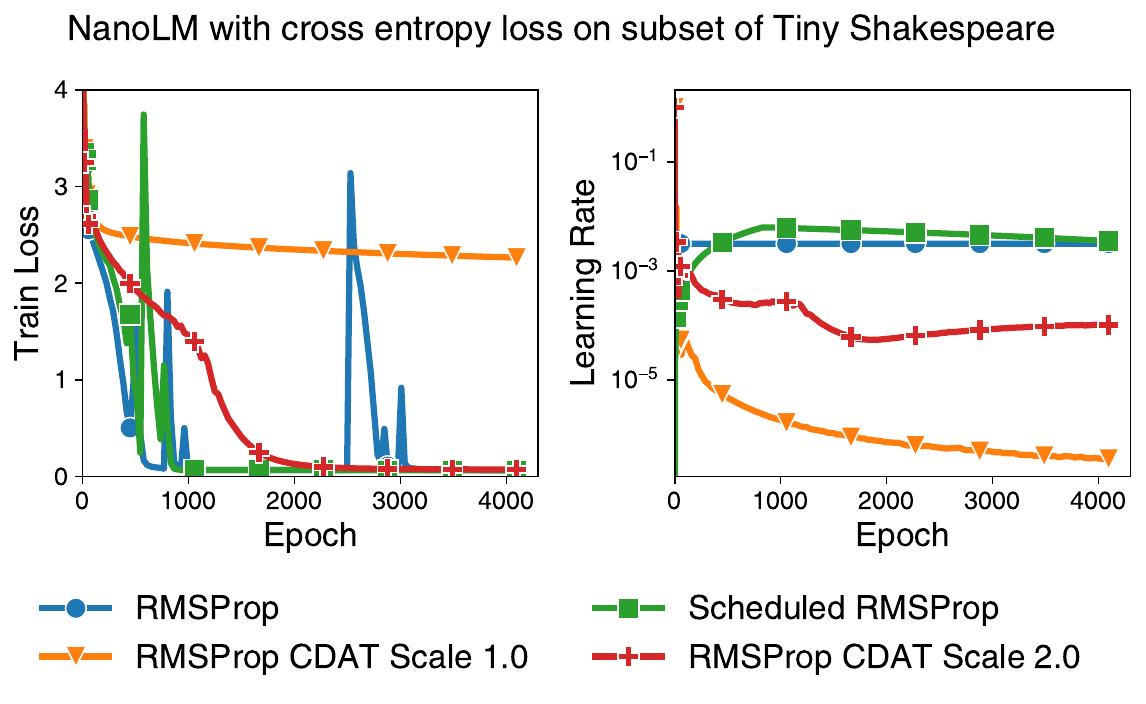}
    \includegraphics[width=0.48\linewidth]{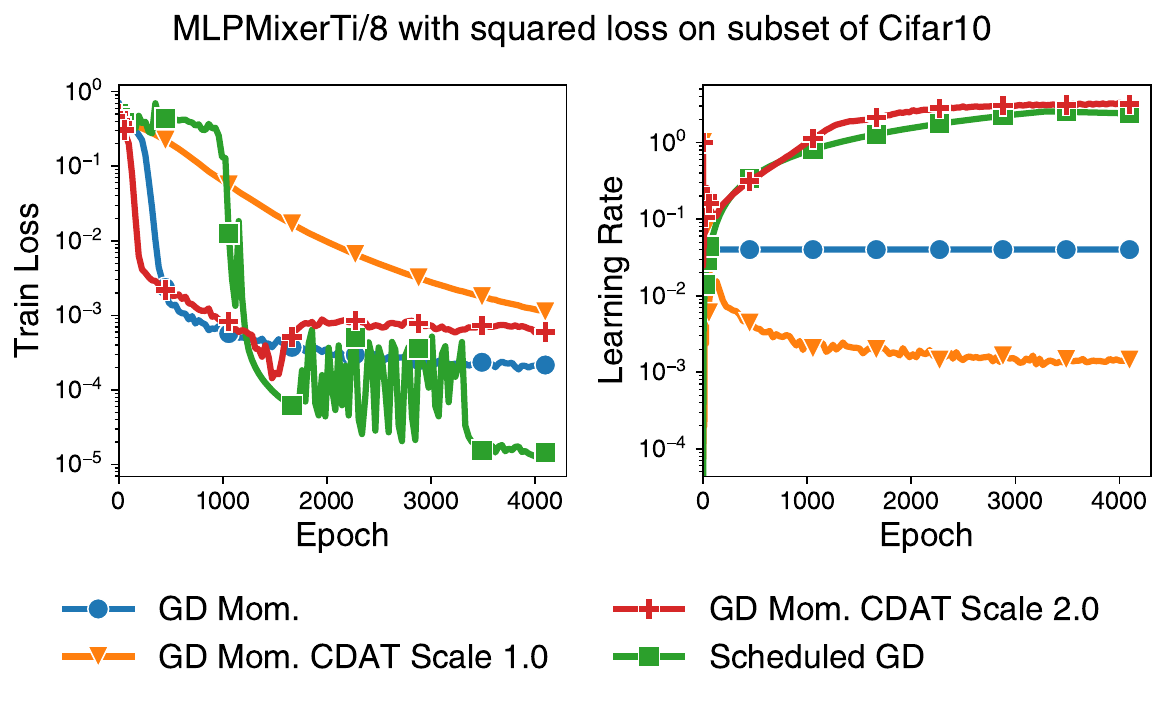}
    \includegraphics[width=0.48\linewidth]{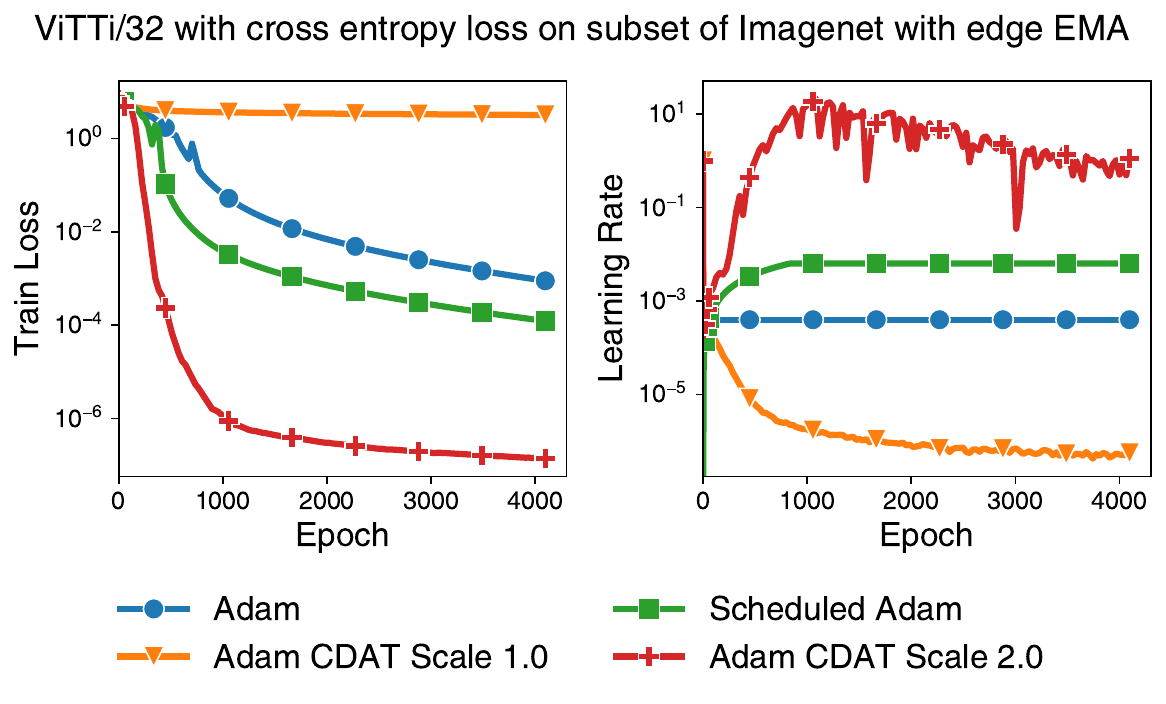}
    \caption{\textbf{CDAT rule may not fully capture the benefits of pre-fixed
    schedules}. Train loss and learning rate behaviors for fine-tuned optimizers
    with or without schedules vs self-tuned counterparts with CDAT on various
    architecture, datasets, losses in a full batch regime. While the CDAT rule
    displays a behavior to warmup schedules, it does not completely catch the
    benefits of pre-fixed schedules. Note for the top-left part that the
    performance of the pre-fixed schedule is akin to the performance reported
    with CDAT $\sigma=2.5$ at early times suggesting that a varying scaling
    factor, or taking higher order dynamics may be important to fully capture
    the benefits of warm-up schedules.}
    \label{fig:on_edge_vs_schedules}
    \vspace*{-15pt}
\end{figure}

In \cref{fig:on_edge_vs_schedules}, we compare the proposed CDAT rule with
prefixed schedules in a full batch regime. We observe that while placing the
optimizer on edge could improve on constant learning rate counterparts, prefixed
schedules can outperform the CDAT rule. This points out that the feedback loop
exploited by CDAT may miss some additional nonlinear effects that could further
enhance self-tuning rules.

\subsection{Detailed performances of CDAT in stochastic regime}
\begin{figure}[t]
    \centering
    \includegraphics[height=0.9\textheight]{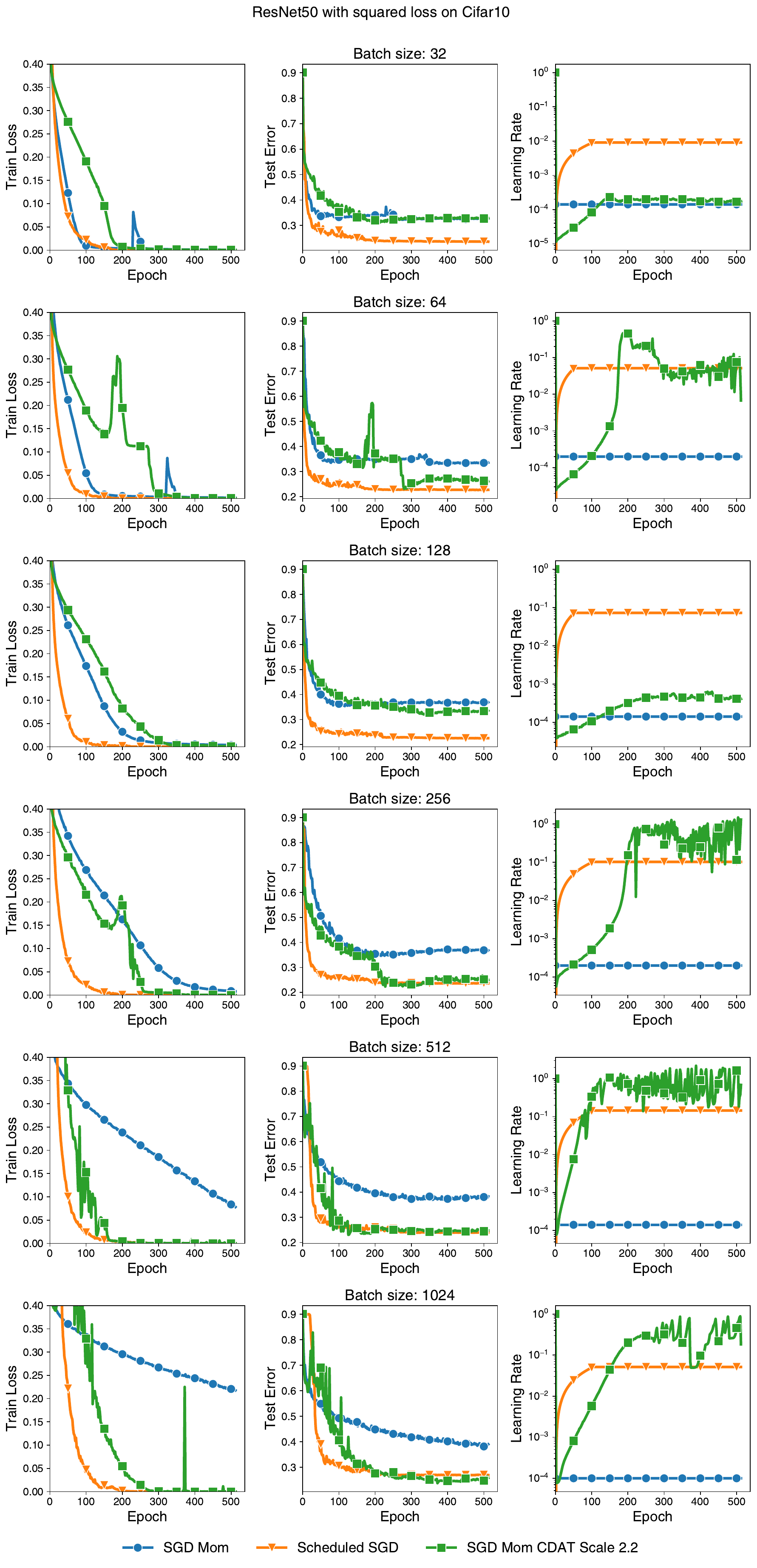}
    \caption{\textbf{Performance of fine-tuned algorithms in stochastic regime
    with SGD Momentum.} In the stochastic regime with SGD momentum, we observe
    that the CDAT rule may outperform the constant learning rate counterpart
    (particularly for large batch sizes) while performing on par or
    underperforming the scheduled learning rate counterparts. Interestingly, a
    warm-up phase appears naturally induced by the CDAT rule (right plots). }
    \label{fig:resnet50_stoch_all_bs}
\end{figure}

\begin{figure}[t]
    \centering
    \includegraphics[width=0.75\linewidth]{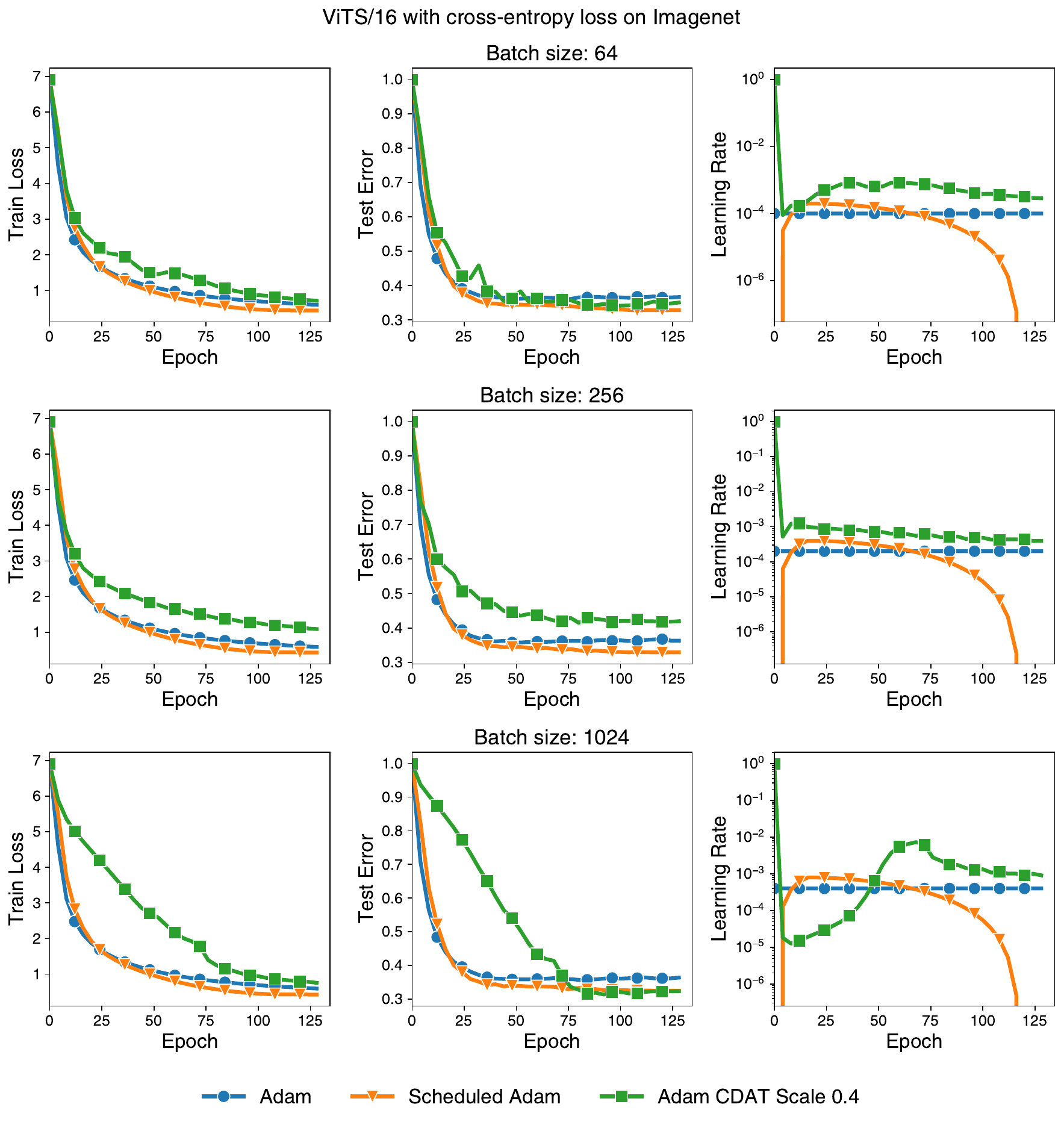}
    \caption{
    \textbf{Performance of fine-tuned algorithms in stochastic regime with Adam.}
    In the stochastic regime with Adam we observe varying performance of the
    CDAT rule, generally on par or underperforming the baselines. Several
    factors can explain the underperformance. The scaling factor is not well
    understood, and a finer grid search could improve the scheme. Similarly, a
    better estimate or a better understanding of the edge of stability in a
    stochastic regime could improve the approach. Tackling curvature dynamics by
    analyzing feedback effects as in the CDAT rule may help such design. }
    \label{fig:vit_adam_stoch_all_bs}
\end{figure}

In~\cref{fig:resnet50_stoch_all_bs} and~\cref{fig:vit_adam_stoch_all_bs}, we
detail the performances of the CDAT rule in the stochastic regime for varying
batch sizes. In the stochastic regime, recall that the heatmap of the
performance of CDAT in terms of batch-size heavily depended on the appropriate
scaling factor~\cref{fig:scaling_factor_vs_batch_size} (in comparison a scaling
factor of approximately $\scale=2$ appeared generally good in the full batch
regime). In both~\cref{fig:resnet50_stoch_all_bs}
and~\cref{fig:vit_adam_stoch_all_bs}, we observe that the method may generally
work better at larger batch sizes. Understanding better the right statistics to
estimate as well as appropriate estimators of the edge of stability in a
stochastic regime is a future direction.

\begin{figure}[t]
    \centering
    \includegraphics[width=\linewidth]{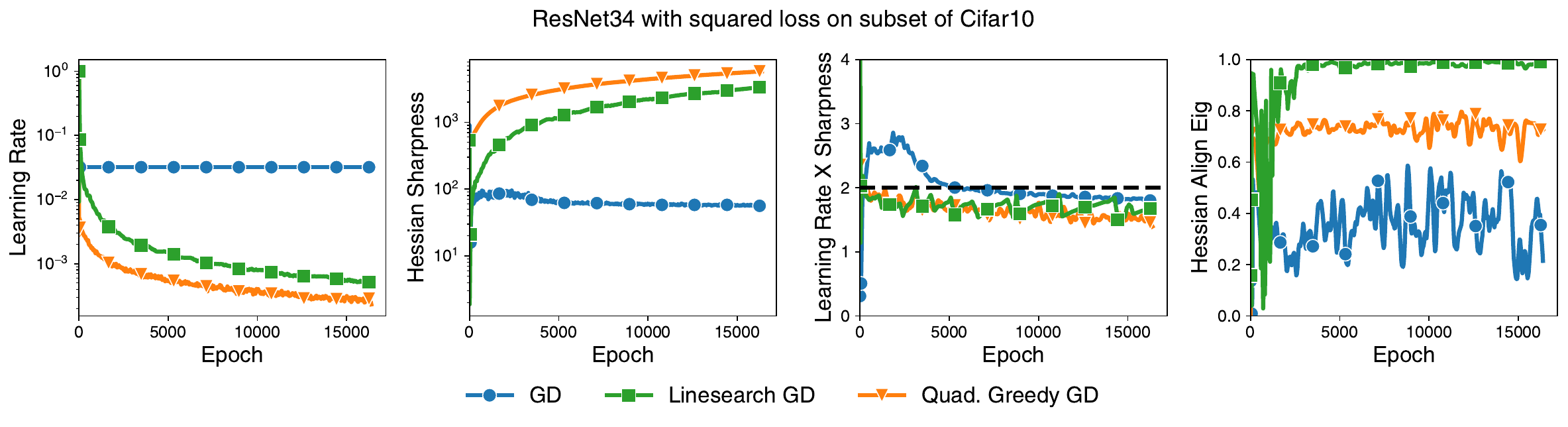}
    \includegraphics[width=\linewidth]{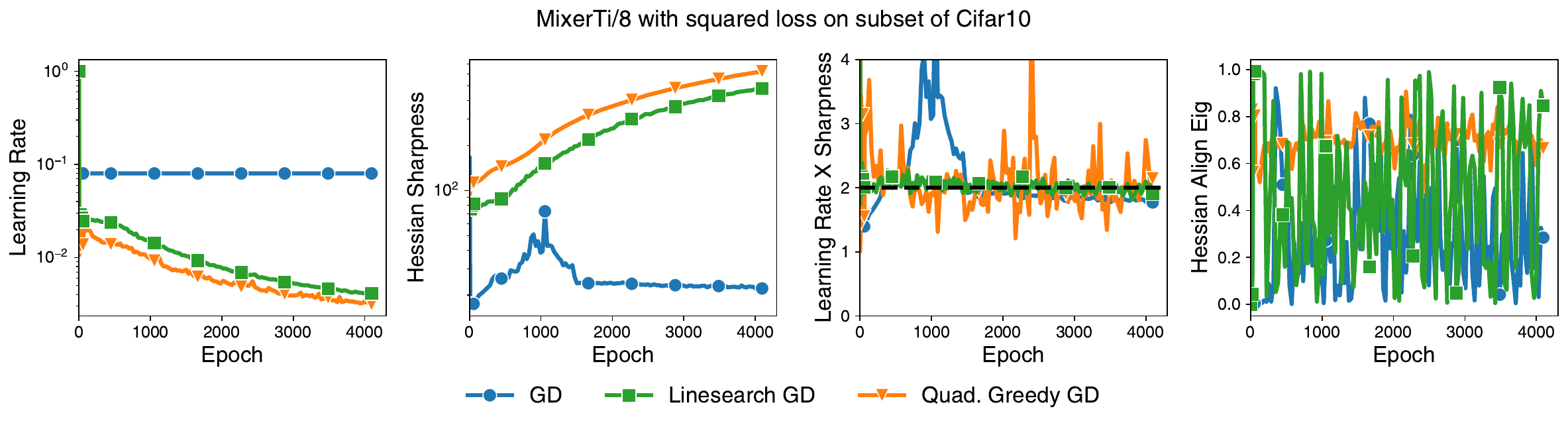}
    \includegraphics[width=\linewidth]{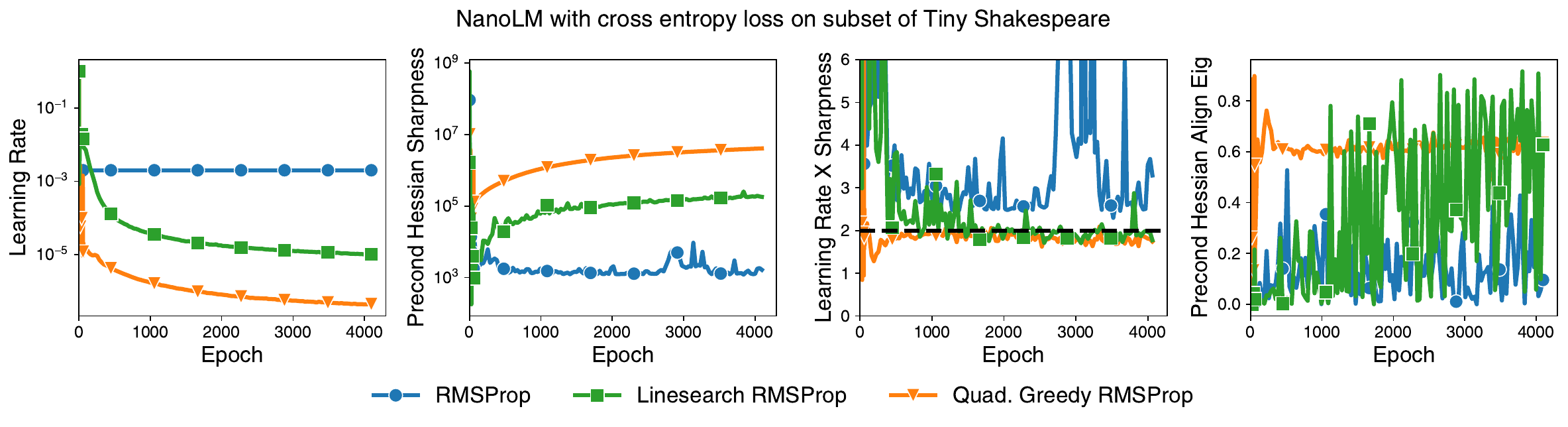}
    \includegraphics[width=\linewidth]{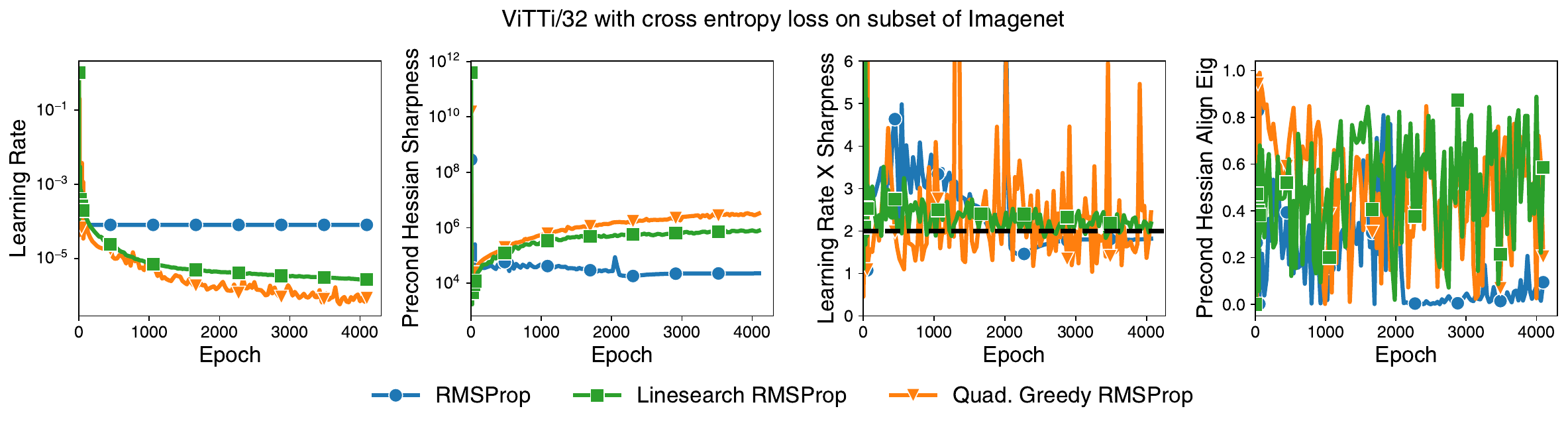}
    \caption{\textbf{Learning rates and sharpness dynamics of baseline learning rate tuners.}
    Learning rate, Hessian or preconditioned Hessian sharpness, their product,
    and the alignment between the update and the largest eigenvector of the
    Hessian or the preconditioned Hessian. As in~\cref{fig:app_sharp_failures},
    we observe that a linesearch~\eqref{eq:linesearch} or a quadratically
    greedy~\eqref{eq:quadratically_greedy} learning rate tuner display
    decreasing learning rates along training. The sharpness of the Hessian (for
    GD) or preconditioned Hessian (for RMSProp) keep increasing for the
    self-tuned baselines while they stabilize for the constant learning rate
    counterparts. The self-tuned methods perform generally below the edge of
    stability or at least much less above than the constant learning rate
    counterpart. }
    \label{fig:app_sharp_failures}
\end{figure}

\begin{figure}[t]
    \centering
    \includegraphics[width=\linewidth]{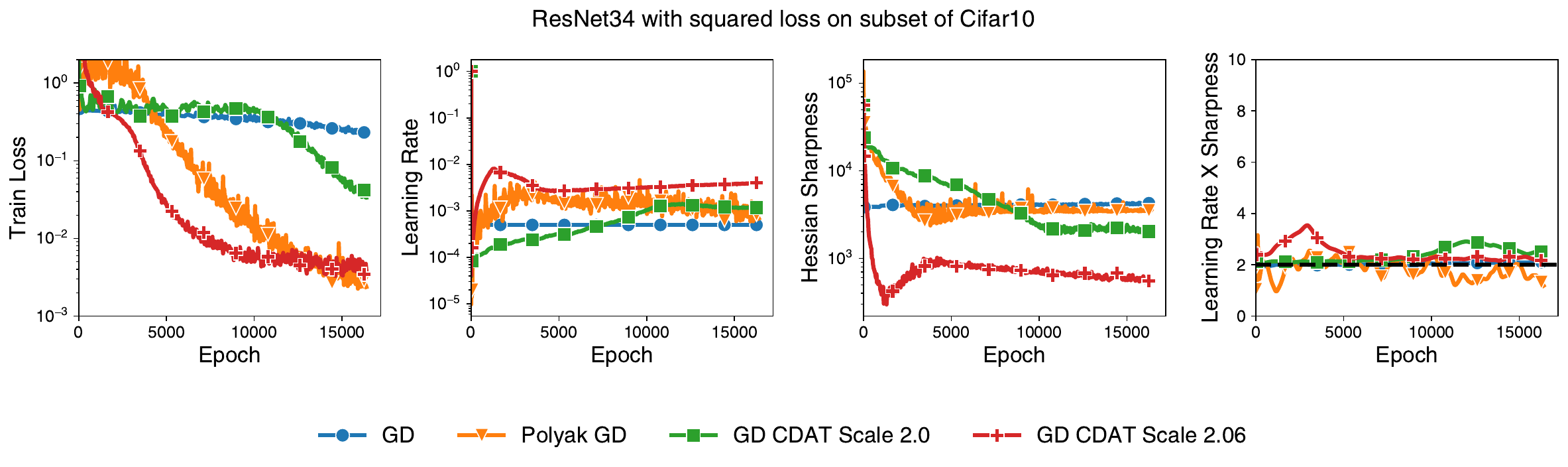}
    \includegraphics[width=\linewidth]{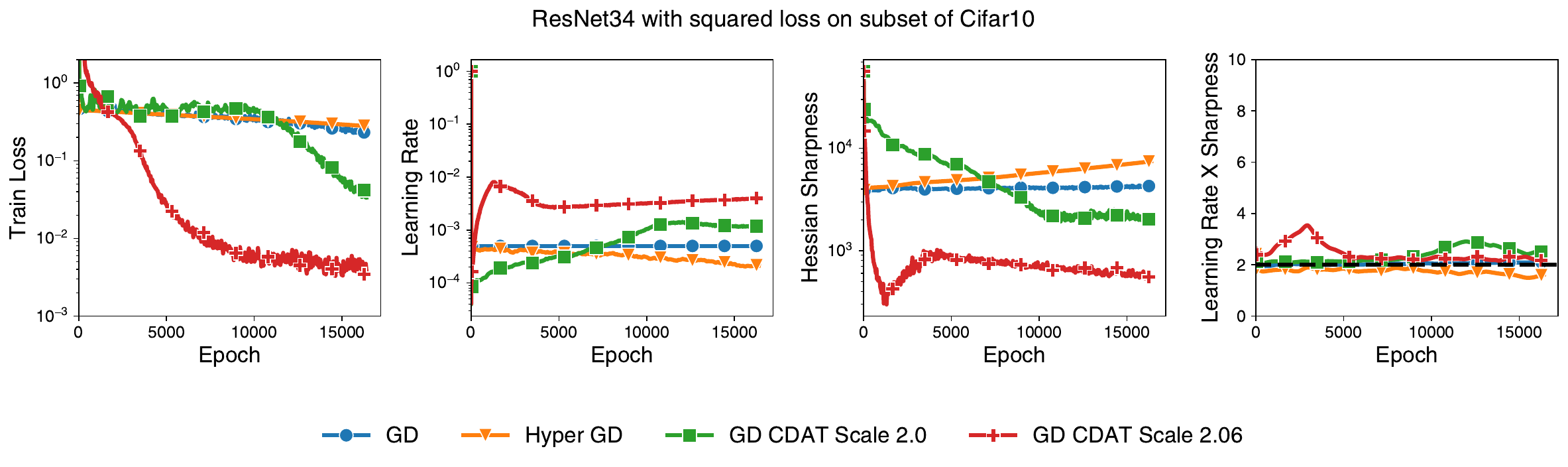}
    \caption{
    \textbf{Analyzing additional learning rate tuners.}
    Train loss, learning rate, sharpness, and their product along training with
    gradient descent using a constant or self-tuned learning rate with various
    tuners. Polyak stepizes~\eqref{eq:polyak} are effective in a full batch
    regime (top figure) outperforming CDAT on edge ($\sigma=2$). The
    effectiveness of the Polyak stepsizes are partially captured by an
    aggressive CDAT rule placing the optimizer on edge $\sigma=2.06$. On the
    other hand, a hyper-gradient descent~\eqref{eq:hypergradient} performs just
    on par with the constant learning rate counterpart in this regime. It also
    displays an ever-increasing sharpening akin to the one observed for a
    linesearch or the quadratically greedy rule~\cref{fig:main_sharp_failures}.
    }
    \label{fig:other_tuners}
\end{figure}

\begin{figure}[t]
    \centering
    \includegraphics[width=0.75\linewidth]{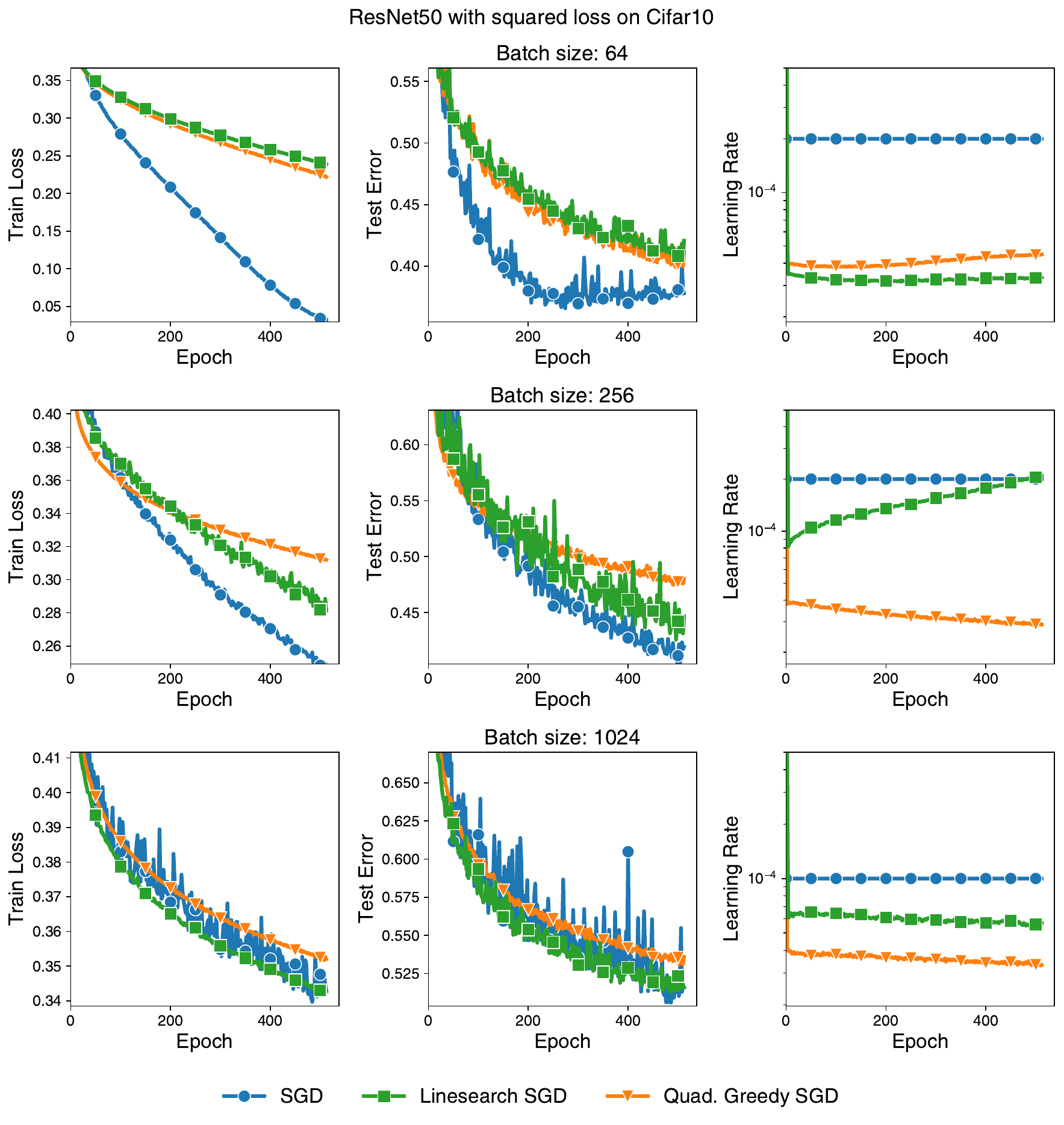}
    \caption{\textbf{Classical learning rate tuners can work well in stochastic regime.}
    In a stochastic regime, we observe that, e.g., a linesearch can perform on
    par or even better than the constant learning counterparts. However, this
    good performance is not explained by the common belief that linesearches
    work well in a full batch regime (\cref{fig:failures_baselines}). The
    linesearch and quadratically greedy rule perform similarly in this setting.}
    \label{fig:baselines_stoch}
\end{figure}

\begin{figure}[t]
    \centering
    \includegraphics[width=0.75\linewidth]{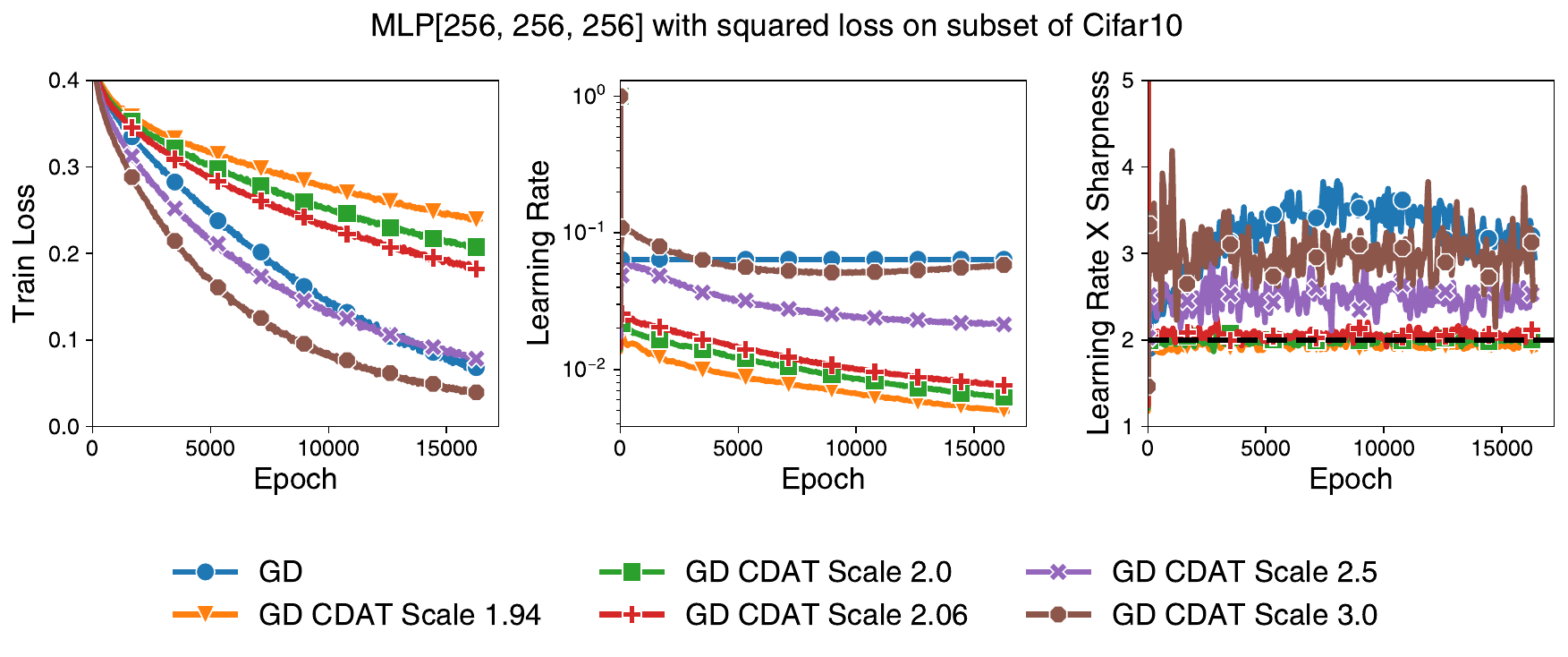}
    \caption{\textbf{On edge with exact sharpness.}
    We consider a rule similar to CDAT~\eqref{eq:gen_on_edge_rule} but using the
    exact sharpness, that is the largest eigenvalue of the Hessian, as a base
    learning rate while varying an additional scale factor
    (see~\eqref{eq:on_exact_edge}). By using the exact sharpness, a scaling
    factor of $\sigma=2$ leads now to poor performance, while a scaling factor
    of $\sigma=3$ is performant. By using the exact
    sharpness~\eqref{eq:on_exact_edge} we do not take into account the actual
    alignment of the update with the largest eigenvector which may explain the
    shift of optimal scaling factors in this case. }
    \label{fig:exact_edge}
\end{figure}

\begin{figure}[t]
    \centering
    \includegraphics[width=0.8\linewidth]{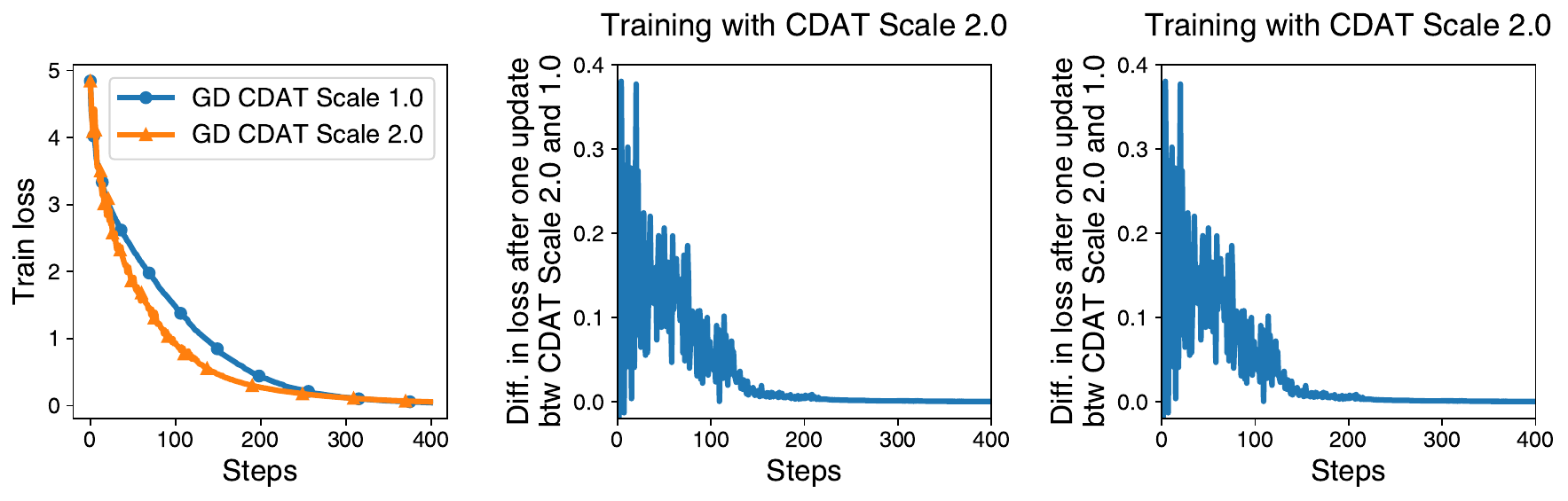}
    \caption{ \textbf{The quadratically greedy rule ensures larger instantaneous
    decrease of the loss.} Train losses and difference in loss between one step
    using the on-edge rule (CDAT, scale=2) and one step using quadratically
    greedy rule (CDAT, scale=1). Results averaged over 10 initializations and
    disjoint 4096-sample subsets of CIFAR100. MLP architecture: single hidden
    layer of size 1024, ReLU activations, trained with GD and cross-entropy loss
    in a full batch setting. We plot $f(w_t + \lr^{\text{oe}}u_t) - f(w_t +
    \lr^{\text{qg}}u_t)$ along a training performed either with the
    quadratically greedy rule ($\scale=1$) or the on-edge rule ($\scale=2$). In
    both cases, this difference is positive meaning that the quadratically
    greedy rule ensures a larger instantaneous decrease of the loss. Yet the
    quadratically greedy rule  underperforms in the long term (see
    \cref{fig:on_edge}, also holds in this full batch setting).
    \label{fig:on_edge_vs_greedy}}
\end{figure}

\begin{figure}[t]
    \centering
    \includegraphics[width=0.6\linewidth]{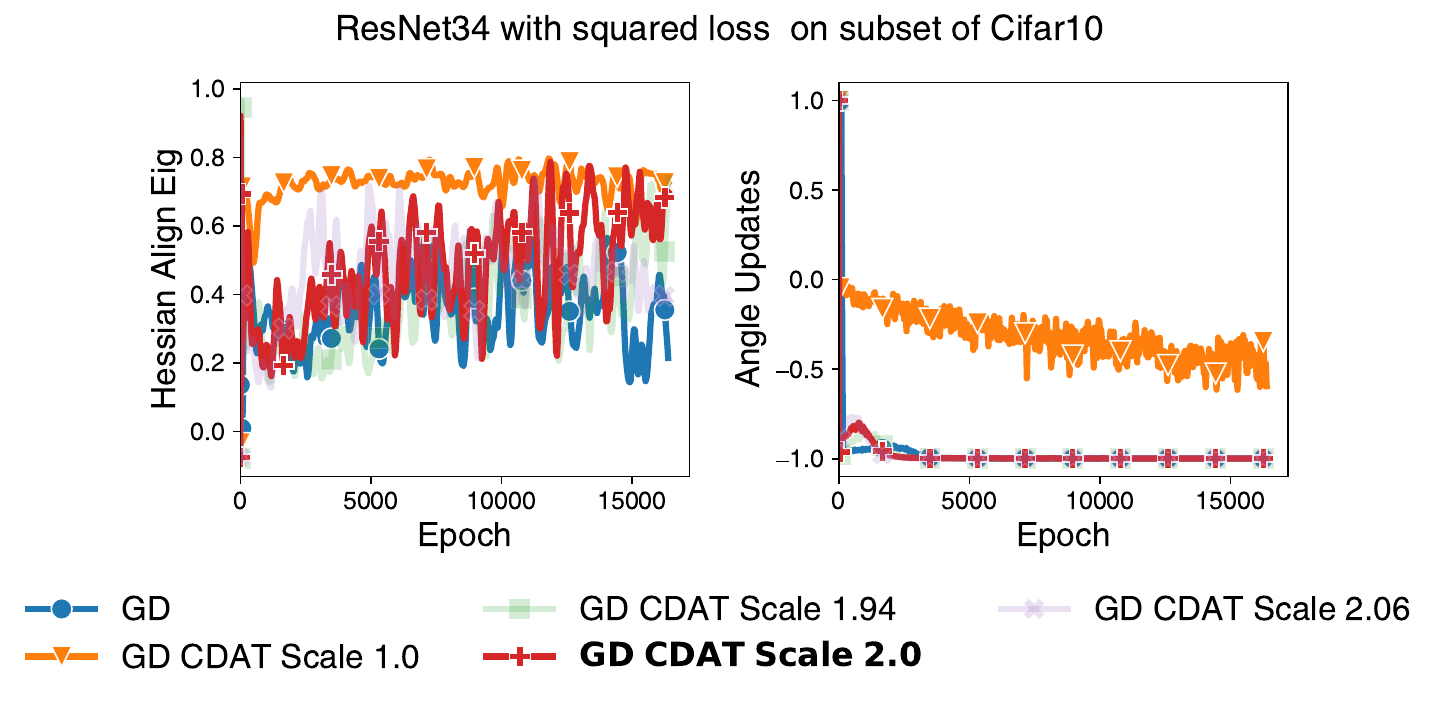}
    \caption{\textbf{Further analysis of the CDAT rule.}
    This is the same setting as in~\cref{fig:on_edge_sharp} except that we plot
    the alignment between the largest eigenvector of the Hessian and the update,
    and the angle between successive updates. We observe that the CDAT rule for
    $\scale \approx 2$ behaves similarly as the constant learning rate
    counterpart. In particular, the updates tend to quickly be in opposed
    directions. The quadratically greedy rule does not demonstrate such a
    behavior.}
    \label{fig:on_edge_more_metrics}
\end{figure}

\begin{figure}[t] 
\centering 
\includegraphics[width=0.35\linewidth]{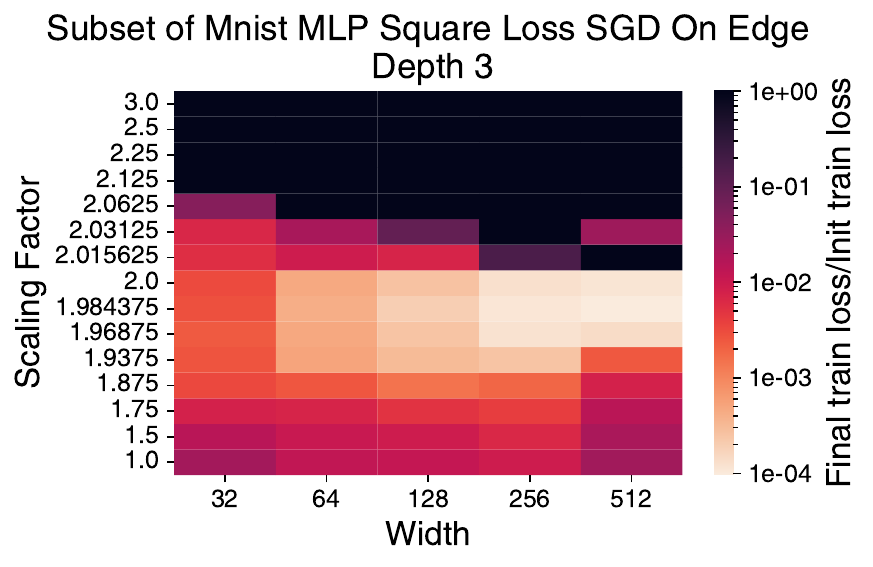}~\hspace*{50pt}
\includegraphics[width=0.35\linewidth]{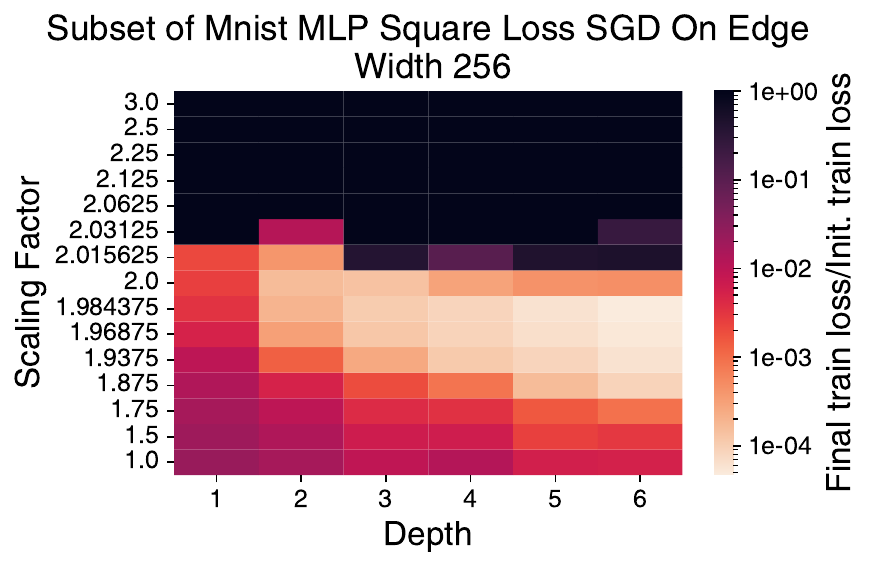}

\includegraphics[width=0.35\linewidth]{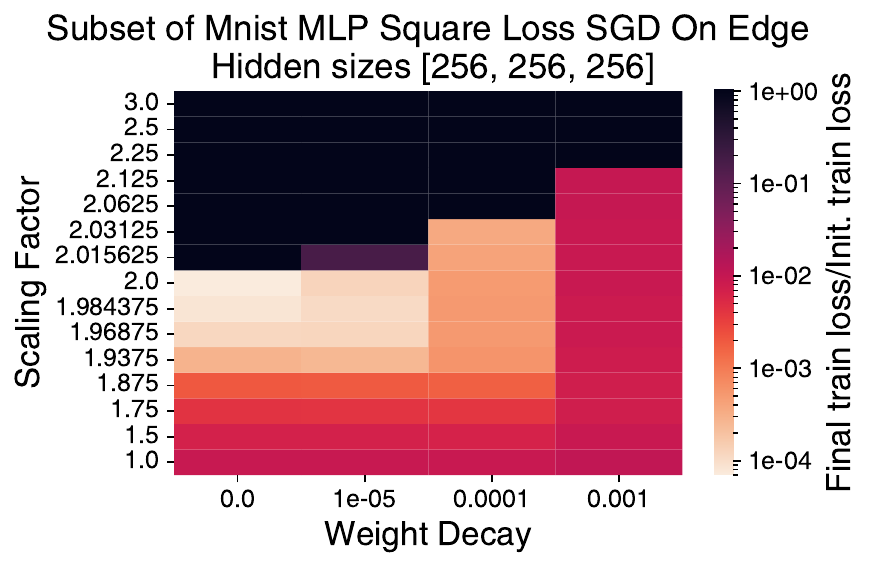}~\hspace*{50pt}
\includegraphics[width=0.35\linewidth]{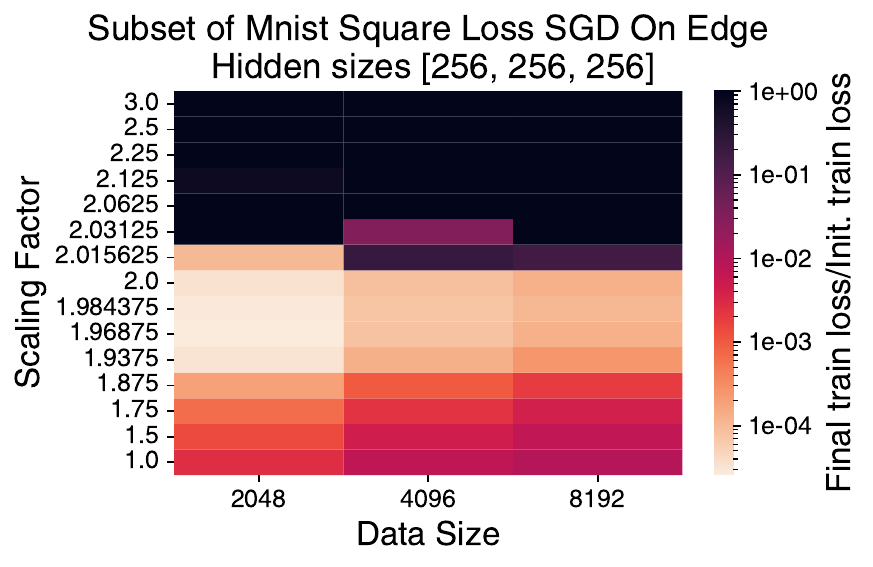}
\caption{\textbf{Improvements of CDAT rule on edge for varying hyperparameters.}
\label{fig:on_edge_mlp_analysis} Varying width, depth, weight decay and size of
the subset considered when using the CDAT rule with varying scaling factors. We
observe that accrued gains can be obtained with the CDAT rule for larger widths
(top left panel). In terms of depth (top right panel), the CDAT rule works best
with larger depths, while we note there a slight shift of optimal scaling
factors from $2$ to slightly below $2$. The CDAT rule appears to work best with
small or no weight decay (bottom left panel) while its benefits fade with larger
weight decay (no difference between greedy $\scale=1$ and on edge $\scale=2$).
Finally, while the method naturally finds smaller train losses with larger
subsets of data, we do not observe a significant shift of relative performance
between scales as the size of the data increases (bottom right panel). }
\end{figure}

\clearpage
\section{Experimental Details}\label{app:exp_details}

\subsection{Datasets}

\paragraph{MNIST} MNIST is an image classification dataset of handwritten
digits~\citep{lecun2010mnist}. We normalized the pictures so that each pixel is
between $0$ and $1$. We did not standardize the data. We only used this dataset
to test varying MLP architectures in~\cref{fig:on_edge_mlp_analysis}.
See~\cref{ssec:details_per_figure} for any additional relevant details.

\paragraph{CIFAR10}
CIFAR10 is an image classification dataset of colored images of size $32\times
32$ with $10$ classes~\citep{krizhevsky2009learning}. We normalized the pictures
so that each pixel is between $0$ and $1$. We did not standardize the data. In
the full batch regime, we considered a subset of $4096$ samples. In the mini
batch regime, we considered the full dataset of $50,000$ samples for training
(dropping out the remainder batch) and tested on the $10,000$ validation
samples.

\paragraph{Tiny Shakespeare} 
This consists in $40,000$ lines of Shakespeare from a variety of Shakespeare's
plays~\citep{karpathy2015rnn}. The task consists in a character-level prediction
task among 64 characters. In the full batch regime, we considered a subset of
$2048$ samples consisting of blocks of $64$ characters.

\paragraph{Imagenet}
Imagenet is an image classification dataset of various images from the
web~\citep{deng2009imagenet}. The images have various sizes. The original
Imagenet-1K dataset contains $1000$ classes. For the full batch experiments, we
consider the Imagenette~\citep{imagenette} subset that consists in only $10$
classes and took $1024$ samples out of it. We consider the usual prepreocessing
for Imagenet as detailed in the Scenic library~\citep{dehghani2021scenic}.
Namely, for training we consider random cropping and random flip at training.
For testing, we center crop the images. Each time the cropping reduces the
colored images to a $224\times 224$ size. In the mini-batch regime we consider
the complete training dataset of $1.2$ million images (Imagenet-1K), dropped the
remainder batch, and reported test error on the $50,000$ validation images.

\subsection{Architectures}

\paragraph{Residual Network (ResNet)}
We considered the standard ResNet architectures (ResNet34, ResNet50)
of~\citet{he2016deep} as implemented in the Scenic
library~\citep{dehghani2021scenic}. For the examples with ResNet34 we removed
the batch normalization layers~\citep{ioffe2015batch}. For the examples with
ResNet50 we replace the batch normalization layers with layer normalization
layers~\citep{ba2016layer}.

\paragraph{Multi-Layer Perceptron (MLP) Mixer}
We consider the standard MLP Mixer architectures~\citep{tolstikhin2021mlp} as
implemented in the Scenic library~\citep{dehghani2021scenic}. By Mixer Ti/8, we
mean the tiny model of Mixer provided in the Scenic library (see
\url{https://github.com/google-research/scenic/blob/main/scenic/projects/baselines/configs/imagenet/imagenet\_augreg\_mixer\_config.py})
with patches of size $8\times 8$. We removed dropout (both layer and depth
wise).

\paragraph{Nano Language Model (NanoLM)}
We consider a simpel sequence-to-sequence Transformer model implemented in Optax
(\url{https://github.com/google-deepmind/optax/blob/main/examples/nanolm.ipynb}).
The model consists of 6 stacked transformer blocks, each of which contains a
multi-head attention layer followed by a feed-forward layer. Layer
normalization is used used within the transformer blocks to improve training
stability.  Finally, a dense layer maps the model's output to the vocabulary
size, producing probabilities for each character as the next potential
character. The only difference with respect to the previous code is that we
removed dropout (for deterministic training) and, as mentioned in the previous
subsection, reduced the size of the dataset to be able to estimate the
sharpness.

\paragraph{Vision Transformer (ViT)}
We consider the standard Vision Transformer
architecture~\citep{dosovitskiy2020image} as implemented in the Scenic
library~\citep{dehghani2021scenic}. By ViT Ti/32, we mean the tiny model of
Vision Transformer provided in the Scenic library (see
\url{https://github.com/google-research/scenic/blob/main/scenic/projects/baselines/configs/imagenet/imagenet_augreg_vit_config.py})
with patches of size $32\times 32$. We removed dropout (both layer and depth
wise).

\paragraph{Weight decay}
In all examples, except mentioned otherwise, we consider a fixed weight decay of
$10^{-5}$. See \cref{fig:on_edge_mlp_analysis} (bottom left panel) for an
analysis of sensitivity of the proposed CDAT rule with varying weight decay.

\subsection{Algorithms}\label{ssec:algo_app}

In all experiments, when selecting the "best" tuner we considered the average
train loss over the last $5$ iterates.

\paragraph{Fine-tuning base optimizers}
The implementation of all optimizers are taken from the Optax
library~\citep{deepmind2020jax}. In all experiments, we fix all hyperparameters
of the base optimizer ((S)GD, (S)GD with Momentum, RMSProp, Adam) to their
default values: $0.9$ for the momentum of (S)GD with Momentum, $0.999$ for the
EMA parameter of the second moment of RMSProp and Adam, $0.9$ for the EMA
parameter of the first moment of Adam. We fine-tune the learning rate on a
logarithmic base of $2$ around a base learning rate such as $10^{-3}$ or
$10^{-4}$ (depending on the algorithm, the architecture and the mini-batch size
in the stochastic regime as detailed below in~\cref{ssec:details_per_figure}),
while making sure that the grid is sufficiently large such that the best
learning rate is found inside the grid and not as the smallest or the largest.

For the scheduled versions of the base optimizers, we consider three shapes:
linear warm-up followed by constant learning rate, linear warm-up followed by
linear decay, linear warm-up followed by cosine decay. The number of iterations
for the warm-up period is chosen as a fraction of the overall number of steps
detailed in~\cref{ssec:details_per_figure}. We also varied the horizon for the
decaying schedules, see again~\cref{ssec:details_per_figure}.

\paragraph{Implementation of the linesearch procedure}
To implement the linesearch procedure described in~\cref{sec:classic_tuners}, we
consider the following criterion
\[
f(\param_t + \lr^\text{ls}_t u_t) 
\leq 
(1+\delta)f(\param_t) + c\lr^{\text{ls}}_t \nabla f(\param_t)^\top u_t.
\]
Compared to~\eqref{eq:linesearch}, we added a relative decrease hyperparmeter
$\delta$ as we observed that the linesearch can sometimes stay stuck at
vanishing learning rates otherwise.

To find a valid criterion we consider a usual backtracking linesearch that
starts from a guess $\lr_{t, 0}=\min\{c_+\lr_{t-1}^{\text{ls}}, 1\}$. Choosing
$c_+ = +\infty$ means that we start with an initial guess of $1$ at each
iteration. The learning rate is then decreased by a factor $c_-$ until the
criterion is satisfied. Formally, the selected stepsize is then
\[
\lr_t^{\text{ls}} 
= \max\{
\lr_{t, k}= c_-^k \lr_{t, 0}:
f(\param_t + \lr_{t, k} u_t)
\leq 
(1+\delta)f(\param_t) + c\lr_{t, k} \nabla f(\param_t)^\top u_t\}.
\]
We run the search until the criterion is satisfied in the full batch regime and
for a maximum of $30$ iterations in the mini-batch regime. In the experiments,
we consider the following variations. 
\begin{itemize}[nosep]
    \item $c\in [0, 10^{-4}, 0.5]$,
    \item $c_+ \in [4, +\infty]$,
    \item $c_- \in [0.8, 0.9]$,
    \item $\delta \in [0, 1e-3]$ for $c=0$ and $\delta = 0$, for $c\in [10^{-4},
    0.5]$.
\end{itemize}

\paragraph{Implementation of quadratically greedy tuner and CDAT}
To implement the quadratically greedy tuner or CDAT, we compute the denominator
$u^\top \nabla^2 f(w) u$ as the second partial derivative of $f$ along $u$, that
is,
\[
u^\top \nabla^2 f(w) u 
= \partial^2 f(w)[u, u] 
= \partial (\partial f(\cdot)[u])(w)[u],
\]
where $\partial g(w)[u]$ amounts to a Jacobian vector product (jvp) computed
with forward mode auto-diff in differentiable programming languages such as
JAX~\citep{deepmind2020jax}.

Computing the denominator in the CDAT rule by forward mode automatic
differentiation enables a much lower memory consumption than using Hessian
vector products (see, e.g., \citep[Chapter 8]{blondel2024elements},
\citep{dagreou2024howtocompute} for more details). The computation of the
denominator by applying twice forward mode automatic differentiation still
incurs approximately three times the memory necessary to compute the
objective~\citep[Chapter 8]{blondel2024elements}. The computational cost of
computing the denominator is also approximately three times the computational
cost of computing the objective. The above approximations are done with the
following reasoning. The second partial derivative requires to follow the graph
of computation but with three variables, one for the parameters, one for a copy
of the update direction, one for another copy. At each node in the computation
graph, the program computes the original computation, computes its first
derivative along the first copy of the update direction, and computes the second
derivative along the second copy.

In practice, we observed for, e.g., the experiment on the full Imagenet dataset
in mini-batch that the proposed CDAT rule required twice the wall time of the
constant or scheduled learning rates counterparts. For this project, we
considered CDAT as a diagnostic tool to understand the interplay between
curvature and learning rate tuners. For future work, the cost of computing the
approximate edge may be circumvented or amortized by using, e.g., parabolic
approximations as done by~\citet{mutschler2020parabolic}, or by computing it at
given intervals as done by~\citet{liu2023sophia}.

\paragraph{Further justification for the CDAT formula}
In~\eqref{eq:gen_on_edge_rule} we took the absolute value of the denominator to
deal with concave approximations. Eigenvalue modifications in a Newton method
are discussed by~\citet[Section 3.4]{nocedal1999numerical}. Taking the absolute
value is one possible option. In practice, we observed positive curvatures along
the update direction such that this choice did not matter.

\subsection{Metrics implementation}

\paragraph{Sharpness estimation}
We estimated the sharpness by a power iteration method run for $1000$ iterations
with an early stopping criterion defined by less than $10^{-3}$ relative
accuracy. We accessed the Hessian by Hessian vector products, which limited the
size of the full batch datasets considered on TPUs. The power iteration a priori
returns the largest eigenvalue in magnitude $|\lambda|_{\max}$ and not
necessarily the largest positive eigenvalue $\lambda_{\max}$. But in practice
the largest eigenvalue in magnitude is the largest eigenvalue, see,
e.g.,~\citep{ghorbani2019investigation} for an in-depth study of the spectrum of
the Hessian along the iterations of deep learning.

\subsection{Additional experimental details per
figure}\label{ssec:details_per_figure} We detail here any additional detail per
figure not detailed in the summary above.

\paragraph{\cref{fig:failures_baselines}} The ResNet34 has no batch
normalization layers. For the GD baseline on ResNet and the MLP Mixer the
constant learning rate was tuned on a grid $\{10^{-3} \cdot 2^i, i \in \{-1,
\ldots, 7\}\}$. For the RMSPRop baseline on the NanoLM and ViT, the constant
learning rate was tuned on a grid $\{10^{-3} \cdot 2^i, i \in \{-1,\ldots 7\}\}$
and $\{10^{-5} \cdot 2^i, i \in \{0, \ldots 7\}\}$ respectively.

\paragraph{\cref{fig:linear_classic_baselines}} We consider a linear
classification (in other words using an MLP without hidden layers) on the subset
of CIFAR10 detailed above. We search the constant stepsize of gradient descent
in $\{10^{-3} \cdot 2^i, i \in \{0, 1, 2\}]\}$. The grid is centered around
$1/\|\nabla^2 f(w_0)\|_2$, that is the optimal stepsize in a convex smooth
setting. As for above experiments, the largest learning rate in the grid led to
divergence.

\paragraph{\cref{fig:main_sharp_failures}} This is the same setting as in the
first panel of \cref{fig:failures_baselines}.

\paragraph{\cref{fig:armijo_theory}} For constant learning rate, model settings
were given by $a = 3\cdot10^{-2}$, $b = 3\cdot10^{-1}$, $\lr = 1$. For fixed $\y
= -0.1$ training, $a = 1$, $b = 0.5$, $\eta_{0} = 1.0$.

\paragraph{\cref{fig:on_edge}} We considered the same settings as
in~\cref{fig:failures_baselines}. The grid search for GD on ResNet34 and RMSProp
on NanoLM are the same as in~\cref{fig:failures_baselines}. For the MLPMixer, we
fine-tuned GD with momentum on a grid $\{10^{-2} \cdot 2^i: i \in \{1, \ldots,
8\}\}$. For the ViT, we fine-tuned Adam on a grid $\{10^{-5} \cdot 2^i: i \in
\{1, \ldots, 8\}\}$.

\paragraph{\cref{fig:on_edge_sharp}} This is exactly the same setting as in the
first panel of \cref{fig:on_edge}.

\paragraph{\cref{fig:scaling_factor_vs_batch_size}} We considered the full
dataset of CIFAR10 with ResNet50 with layer normalization in place of batch
normalization.

\paragraph{\cref{fig:on_edge_stoch}} See the details given for
\cref{fig:resnet50_stoch_all_bs} and \cref{fig:vit_adam_stoch_all_bs}.

\paragraph{\cref{fig:edge_theory}} For both values of $\scale$, $a =
5\cdot10^{-2}$, $b = 10^{-1}$, $\nu = 0.1$, $\lam_{0} = 18$, $\eta_{0} = 0.05$,
$g_{0} = \p_{0} = 4$.

\paragraph{\cref{fig:p_dynamics_armijo}} Same settings as Figure
\cref{fig:armijo_theory}.

\paragraph{\cref{fig:p_dynamics_on_edge}} Same settings as Figure
\cref{fig:edge_theory}.

\paragraph{\cref{fig:app_sharp_failures}} This is the same setting as
in~\cref{fig:failures_baselines}.

\paragraph{\cref{fig:other_tuners}} We consider the same setting as
in~\cref{fig:main_sharp_failures}. For the Polyak stepsizes~\eqref{eq:polyak},
we let $\lr_{\max}$ vary between $1$ and $100$ and select the best. For the
hypergradient descent, we let the hyper learning rate vary in $\beta \in \{10^i,
i \in \{-3, \ldots,  0\}\}$.

\paragraph{\cref{fig:baselines_stoch}} We considered again the ResNet50 with
layer normalization instead of batch normalization. For the constant learning
rate baseline we searched over a grid of $\{\lr_m \cdot 2^i, i \in \{-1, \ldots,
5\}]\}$, for $\lr_m = 10^{-4} \cdot \sqrt{m/4096}$ for $m$ the batch size.

\paragraph{\cref{fig:exact_edge}} We considered a simple MLP with hidden sizes
$(256, 256, 256)$, ReLU activations. We tuned the constant learning rate
baseline on $\{10^{-3} \cdot 2^i, i \in \{-1, 7\}]\}$.

\paragraph{\cref{fig:on_edge_vs_greedy}} Details are provided in the legend.

\paragraph{\cref{fig:on_edge_more_metrics}} This is the same setting as
in~\cref{fig:on_edge_sharp}.

\paragraph{\cref{fig:on_edge_mlp_analysis}} In this figure, the MLPs considered
use ReLu activations. If not detailed, the weight decay is set to $10^{-5}$ and
the subset considered is of size $8192$.

\paragraph{\cref{fig:on_edge_vs_schedules}} The settings are the same as in
\cref{fig:on_edge}. For the constant learning rate baselines we searched on a
gird $\{\lr_{\text{base}}\cdot 2^i, i \in \{-3, \ldots, 9\}\}$. The base
learning rate $\lr_{\text{base}}$ was chosen to be $10^{-3}$ for ResNet,
$10^{-2}$ for the Mixer, $10^{-4}$ for the NanoLM and ViT. For the schedules'
shapes, we searched over linear warm-up, linear warm-up with linear decay,
linear warm-up with cosine decay. The initial and end learning rate were set to
$0$. The horizons for the schedules were chosen in $[N, N/2, N/4]$ for $N=8192$
for the NanoLM, $N=16384$ for the ViT, Mixer and ResNet. The fraction of warm-up
steps was searched in $\{0.05, 0.1, 0.2\}$.

\paragraph{\cref{fig:resnet50_stoch_all_bs}} For the constant learning rate
baseline, we consider searching the best constant learning rate on a grid
$\{\lr_{m} \cdot 2^i, i \in \{-1, \ldots, 7\}\}$ for $\lr_{m} = 10^{-4} \cdot
\sqrt{m/4096}$ where $m$ denotes the varying batch size.

For the scheduled baseline, we consider the variants presented above (linear
warm-up followed by constant, linear warm-up followed by cosine decay, linear
warm-up followed by linear decay) with varying fraction of warm-up steps ($0.05,
0.1, 0.2$) and an initial learning rate of $0$, a final learning rate of $0$ for
a fixed horizon of $512$ epochs, and a peak learning rate searched over
$\{\lr_{m} \cdot 4^i, i \in \{4, \ldots, 9\}\}$.

The scaling factor $\scale$ of CDAT was searched on a grid $\{0.4, 0.6, \ldots,
2.8\}$, and we also tuned the EMA parameter $\beta_{\text{cdat}}$ in the
computation of the numerators and denominators of the edge in $\{0, 0.9,
0.99\}$. The best parameters found for CDAT can be inferred
from~\cref{fig:scaling_factor_vs_batch_size}. Namely, we found that non-zero EMA
parameter for the estimation of the edge decay was essential for good
performance and that the best scaling factor varied with the batch size. For
example, at batch size $256$ the best scaling factor is $\sigma =1.8$ with
$\beta_{\text{cdat}}=0.9$.

\paragraph{\cref{fig:vit_adam_stoch_all_bs}} For the constant learning rate
baseline, we consider searching the best constant learning rate on a grid
$\{\lr_{m} \cdot 2^i, i \in \{-1, \ldots, 7\}\}$ for $\lr_{m} = 10^{-4} \cdot
\sqrt{m/1024}$ where $m$ denotes the varying batch size.

For the scheduled baseline, we consider a linear warm-up followed by cosine
decay, with a fraction of warm-up steps of $0.1$ and an initial learning rate of
$0$, a final learning rate of $0$ for a fixed horizon of $128$ epochs, and a
peak learning rate searched over $\{\lr_{m} \cdot 4^i, i \in \{1, \ldots,
5\}\}$.

The scaling factor $\scale$ of CDAT was searched on a grid $\{0.4, 0.6, \ldots,
2.6\}$, and we also tuned the EMA parameter $\beta_{\text{cdat}}$ in the
computation of the numerators and denominators of the edge in $\{0, 0.9,
0.99\}$.

\subsection{Assets license and computing ressources}\label{ssec:ressources}
\paragraph{Assets}
All experiments are done in the open-source JAX
ecosystem~\citep{deepmind2020jax}: architectures are taken from
Scenic~\citep{dehghani2021scenic}, datasets from TensorFlow Dataset, algorithms
from Optax. The datasets are MNIST~\citep{lecun2010mnist}, (Creative Commons
Attribution-Share Alike 3.0 license) CIFAR10~\citep{krizhevsky2009learning} (no
available license), Imagenet~\citep{deng2009imagenet} (ImageNet explicitly
permits the use of the dataset for non-commercial research purposes, however
there is no single license since the images are scrapped from different sources
with different licenses), TinyShakespeare~\citep{karpathy2015rnn} (Apache 2.0
license in TensorFlow dataset, though the works of William Shakespeare are in the
public domain).

\paragraph{Computing resources}
Experiments have mostly been run on Tensor Processing Units (TPUs) v2 (180 Tera
Floating-Point Operations per Second (TFLOPS), 64 GB High Bandwidth Memory
(HBM)). Experiments on MLP Mixers required TPUs v3 (420 TFLOPS 128 GB HBM). Very
small scale experiments on MNIST with MLPs were run on CPUs. In terms of wall
time, as discussed in \cref{ssec:algo_app}, we observed that the CDAT rule can
be twice slower than the constant or scheduled learning rate counterparts. We
consider CDAT as a diagnostic tool and leave as future work efficient
implementations. Preliminary experiments and additional attempts to further
adapt the momentum parameter on edge are not reported.

\section*{Authors contributions}

\begin{itemize}[nosep, leftmargin=*]
  \item Vincent Roulet conducted the experimental work from the failures of
  linesearches to the analysis of the CDAT rule in various settings.
  \item Atish Agarwala developed the theoretical model, with associated
  figures, interpretations and comments. He also helped to guide the
  experimental study with the insights gathered by the model. 
  \item Jean Bastien Grill did an initial empirical study that gathered first
  intuitions on the method. He participated in the discussions and contributed
  to the writing.
  \item Grzegorz Swirszcz participated in the discussions.
  \item Mathieu Blondel participated in the discussions, contributed to the writing
  and proposed an alternative rule using a Gauss-Newton approximation of the objective.
  \item Fabian Pedregosa initiated the project, performed a larger scale
  empirical study of the CDAT rule on the MLCommons benchmark, participated in
  the discussions, and contributed to the writing.
\end{itemize}
\end{document}